%% file: Formatting-Instructions-LaTeX-2025.tex
\documentclass[letterpaper]{article} % DO NOT CHANGE THIS
\usepackage{aaai25}  % DO NOT CHANGE THIS
\usepackage{times}  % DO NOT CHANGE THIS
\usepackage{helvet}  % DO NOT CHANGE THIS
\usepackage{courier}  % DO NOT CHANGE THIS
\usepackage[hyphens]{url}  % DO NOT CHANGE THIS
\usepackage{graphicx} % DO NOT CHANGE THIS
\usepackage{natbib}  % DO NOT CHANGE THIS AND DO NOT ADD ANY OPTIONS TO IT
\usepackage{caption} % DO NOT CHANGE THIS AND DO NOT ADD ANY OPTIONS TO IT
\usepackage{lipsum}
\usepackage{algorithm2e}

\RestyleAlgo{ruled}
\SetKwComment{Comment}{$\textcolor{gray}{\triangleright}$\ }{}

\newcommand{\topic}[1]{\vspace{0.0in}\noindent\textbf{#1.}}

\SetKwInOut{Input}{Input}
\SetKwInOut{Output}{Output}
\SetKwInOut{Require}{Require}

\usepackage{amsmath}
\newtheorem{theorem}{Theorem}
\input{math_commands.tex}

\usepackage{booktabs}
\usepackage{xcolor}
\usepackage{multirow}

\usepackage{newfloat}
\usepackage{listings}
\DeclareCaptionStyle{ruled}{labelfont=normalfont,labelsep=colon,strut=off} % DO NOT CHANGE THIS
\title{Can Watermarking Large Language Models Prevent \\ Copyrighted Text Generation and Hide Training Data?}
\author{
    Michael-Andrei Panaitescu-Liess$^{\S}$, Zora Che$^{\S}$, Bang An$^{\S}$, Yuancheng Xu$^{\S}$, Pankayaraj Pathmanathan$^{\S}$, Souradip Chakraborty$^{\S}$, Sicheng Zhu$^{\S}$, Tom Goldstein$^{\S}$, Furong Huang$^{\S, \P}$
}
\affiliations{
    $^{\S}$University of Maryland, College Park  \hspace{30px}   $^{\P}$Capital One\\
    \{mpanaite, zche, bangan, ycxu, pan, schakra3, sczhu, tomg, furongh\}@umd.edu
}

\usepackage{bibentry}
\begin{document}

\maketitle

\input{sections/abstract}

\input{sections/introduction}

\input{sections/related}

\input{sections/background}

\input{sections/verbatim}

\input{sections/mia}

\input{sections/method}

\input{sections/conclusion}

\input{sections/ack}

\bibliography{aaai25}

\input{sections/appendix}

\end{document}

%% file: math_commands.tex
\usepackage{amsmath,amsfonts,bm}

\def\eqref#1{equation~\ref{#1}}
\def\1{\bm{1}}

\DeclareMathAlphabet{\mathsfit}{\encodingdefault}{\sfdefault}{m}{sl}
\SetMathAlphabet{\mathsfit}{bold}{\encodingdefault}{\sfdefault}{bx}{n}

%% file: sections/abstract.tex
\begin{abstract}
Large Language Models (LLMs) have demonstrated impressive capabilities in generating diverse and contextually rich text. However, concerns regarding copyright infringement arise as LLMs may inadvertently produce copyrighted material. In this paper, we first investigate the effectiveness of watermarking LLMs as a deterrent against the generation of copyrighted texts. Through theoretical analysis and empirical evaluation, we demonstrate that incorporating watermarks into LLMs significantly reduces the likelihood of generating copyrighted content, thereby addressing a critical concern in the deployment of LLMs. However, we also find that watermarking can have unintended consequences on Membership Inference Attacks (MIAs), which aim to discern whether a sample was part of the pretraining dataset and may be used to detect copyright violations. Surprisingly, we find that watermarking adversely affects the success rate of MIAs, complicating the task of detecting copyrighted text in the pretraining dataset. These results reveal the complex interplay between different regulatory measures, which may impact each other in unforeseen ways. Finally, we propose an adaptive technique to improve the success rate of a recent MIA under watermarking. Our findings underscore the importance of developing adaptive methods to study critical problems in LLMs with potential legal implications.
\end{abstract}

%% file: sections/introduction.tex
\section{Introduction}

\begin{figure*}
    \centering
    \includegraphics[width=0.7\textwidth]{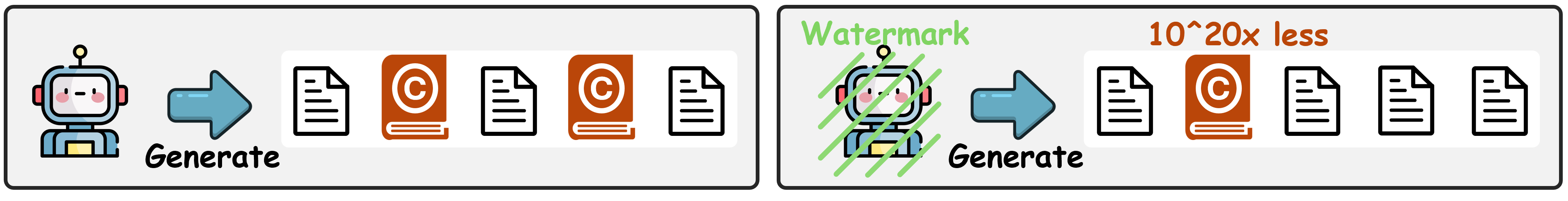}
    \caption{Illustration of the effect of LLM watermarking on generation of copyrighted content. We observe that a moderately strong watermark ($\delta$ = 10) can make it more than $10^{20}$ times less likely for Llama-30B to generate copyrighted content.}
    \label{fig:intro-diagram}
\end{figure*}

In recent years, Large Language Models (LLMs) have pushed the frontiers of natural language processing by facilitating sophisticated tasks like text generation, translation, and summarization. With their impressive performance, LLMs are increasingly integrated into various applications, including virtual assistants, chatbots, content generation, and education. However, the widespread usage of LLMs brings forth serious concerns regarding potential copyright infringements. Addressing these challenges is critical for the ethical and legal deployment of LLMs. \\ \\
Copyright infringement involves unauthorized \text{usage of} copyrighted content, which violates the intellectual property rights of copyright owners, potentially undermining content creators' ability to fund their work, and affecting the diversity of creative outputs in society. Additionally, violators can face legal consequences, including lawsuits and financial penalties. For LLMs, copyright infringement can occur through (1) generation of copyrighted content during deployment and (2) illegal usage of copyrighted works during training. Ensuring the absence of copyrighted content in the vast training datasets of LLMs is challenging. Moreover, legal debates around generative AI copyright infringement vary by region, complicating compliance further. \\ \\
Current lawsuits against AI companies for unauthorized use of copyrighted content (e.g., Andersen v. Stability AI Ltd, NYT v. OpenAI) highlight the urgent need for methods to address these challenges. In this paper, we focus on studying the effects of watermarking LLMs on two critical issues: (1) preventing the generation of copyrighted content, and (2) detecting copyrighted content in training data. We show that watermarking can significantly impact both the generation of copyrighted text and the detection of copyrighted content in training data. \\ \\
Firstly, we observe that current LLM output watermarking techniques can significantly reduce the probability of LLMs generating copyrighted content, by tens of orders of magnitude. Our empirical results focus on two recent watermarking methods: UMD~\citep{kirchenbauer2023watermark} and Unigram-Watermark~\citep{zhao2023provable}. Both methods split the vocabulary into two sets (green and red) and bias the model towards selecting tokens from the green set by altering the logits distribution, thereby embedding a detectable signal. We provide both empirical and theoretical results to support our findings. \\ \\
Secondly, we demonstrate that watermarking techniques can decrease the success rate of Membership Inference Attacks (MIAs), which aim to detect whether a piece of copyrighted text was part of the training dataset. Since MIAs exploit the model’s output, their performance can suffer under watermarking due to changes in the probability distribution of output tokens. Our comprehensive empirical study, including 5 recent MIAs and 5 LLMs, shows that the AUC of detection methods can be reduced by up to $16\%$ in the presence of watermarks. \\ \\
Finally, we propose an adaptive method designed to enhance the success rate of a recent MIA~\citep{shi2023detecting} in detecting copyright violations under watermarking. This method applies a correction to the model’s output to account for the perturbations introduced by watermarks. By incorporating knowledge about the watermarking scheme, we improve the detection performance for pretraining data, counteracting the obfuscation caused by watermarking. Our contribution underscores the importance of continuously developing adaptive attack methodologies to keep pace with advances in defense mechanisms. \\ \\
The rest of the paper is organized as follows. In the ``Related Work'' section, we review prior research on LLM watermarking and copyright. The ``Setup and Notations'' section formally introduces the problems we study. We then present our first two contributions and introduce the adaptive version of the \text{Min-K\%} Prob membership inference attack in the following three sections. Finally, we provide concluding remarks in the last section. Additional experiments, theoretical results, and a discussion on the limitations of our work are included in the appendix
% ~\footnote{The appendix is available in the arXiv version of the paper (https://arxiv.org/abs/2407.17417).}.

%% file: sections/related.tex
\section{Related Work}

\label{related}

\topic{Watermarks for LLMs}
Language model watermarking techniques embed identifiable markers into output text to detect AI-generated content.
Recent strategies incorporate watermarks during the decoding phase of language models~\citep{zhao2023provable,kirchenbauer2023watermark}. \citet{aaronson2023watermark} develops the Gumbel watermark, which employs traceable pseudo-random sampling for generating subsequent tokens. \citet{kirchenbauer2023watermark} splits the vocabulary into red and green lists according to preceding tokens, biasing the generation towards green tokens. 
\citet{zhao2023provable} employs a % simplified 
fixed grouping strategy to develop a robust watermark with theoretical guarantees.  
\citet{liu2024a} proposes to generate watermark logits based on the preceding tokens’ semantics rather than their token IDs to boost the robustness. 
\citet{kuditipudi2023robust} and \citet{christ2023undetectable} explore watermark methods that do not change the output textual distribution. \\ \\
\topic{Copyright}
Copyright protection in the age of AI has gained importance, as discussed by~\citet{ren2024copyright}. 
\citet{vyas2023provable} addresses content protection through near access-freeness (NAF) and developed learning algorithms for generative models to ensure compliance under NAF conditions. Prior works focus on training algorithms to prevent copyrighted text generation~\citep{vyas2023provable, chu2024protect}, whereas our work emphasizes lightweight, inference-time algorithms.
Other works have studied copyright in machine learning from a legal perspective.
\citet{hacohen2024not} utilizes a generative model to determine the generic characteristics of works to aid in defining the scope of copyright.
\citet{elkin2023can} demonstrates that copying does not necessarily constitute copyright infringement and argues that existing detection methods may detract from the foundational purposes of copyright law. \\ \\
Additionally, we include a discussion on memorization and membership inference in the appendix.

%% file: sections/background.tex
\section{Setup and Notations}\label{background}

\subsection{Definitions}
Let $D$ be a training dataset, $C$ be all the copyrighted texts, and $C_D$ be all the copyrighted texts that are part of $D$. We give definitions for the following setups. \\ \\
\topic{Verbatim Memorization of Copyrighted Content} For a fixed $k \in \mathbb{N}$, \citet{carlini2022quantifying} defines a string $s$ as being memorized by a model if $s$ is extractable with a prompt $p$ of length $k$ using greedy decoding and the concatenation $p \oplus s \in D$. We adopt a similar definition for verbatim memorization of copyrighted content but employ a continuous metric to measure it. Specifically, we measure verbatim memorization of a text $c \in C$ using the perplexity of the model on the copyrighted text $c_p$ when given the prefix $p$ as a prompt (where $c_p$ represents the text $c$ after removing its prefix $p$). Note that for $c_p = c_p^{(1)} \oplus c_p^{(2)} \oplus \dots \oplus c_p^{(n)}$ we compute the perplexity using the following formula $\texttt{perplexity}(c_p | p) = \big(\prod_{i=1}^{n} \mathbb{P}(c_p^{(i)} | p \oplus c_p^{(0)} \oplus c_p^{(1)} \oplus \dots \oplus c_p^{(i-1)}\big)^{-\frac{1}{n}}$, where $c_p^{(0)}$ is the empty string. In our experiments, $p$ is either an empty string or the first $10$, $20$, or $100$ tokens of $c$. Lower perplexity thereby indicate higher levels of memorization. \\ \\
\topic{MIAs for Copyrighted Training Data Detection} MIAs are privacy attacks aiming to detect whether a sample was part of the training set. We define an MIA for copyrighted data as a binary classifier $A(\cdot)$, which ideally outputs $A(x) = 1, \forall x \in C_D$ and $A(x)= 0, \forall x \in C - C_D$. In practice, $A(\cdot)$ is defined by thresholding a metric (e.g., perplexity), i.e., $A(x) = 1, \forall x$ such that \texttt{perplexity}$(x) < t$ and $0$, otherwise. Since the threshold $t$ needs to be set, prior work~\citep{shi2023detecting} uses \texttt{AUC} (Area Under the ROC Curve) as an evaluation metric which is independent of $t$. Note that we employ the same metric in our experiments. \\ \\
\topic{LLM Watermarking} Watermarking LLMs consists of introducing signals during its training or inference that are difficult to detect by humans without the knowledge of a \textit{watermark key} but can be detected using an algorithm if the key is known. We focus our paper on recent methods that employ logits distribution changes as a way of inserting watermark signals during the decoding process \cite{kirchenbauer2023watermark, zhao2023provable}.

\subsection{MIAs}

Current MIAs for detecting training data rely on thresholding various heuristics that capture differences in output probabilities for each token between data included in the training set and data that was not. Below, we present an overview of these heuristics. \\ \\
\topic{Perplexity} This metric distinguishes between data used to train the model (members) and data that was not (non-members), as members are generally expected to have lower perplexity. \\ \\
\textbf{Smaller Ref, Lowercase and Zlib}~\citep{carlini2021extracting}. Smaller Ref is defined as the ratio of the log-perplexity of the target LLM on a sample to the log-perplexity of a smaller reference LLM on the same sample. Lowercase represents the ratio of the log-perplexity of the target LLM on the original sample to the log-perplexity of the LLM on the lowercase version of the sample. Zlib is defined as the ratio of the log-perplexity of the target LLM on a sample to the zlib entropy of the same sample. \\ \\
\textbf{Min-K\% Prob}~\citep{shi2023detecting}. This heuristic computes the average of the minimum $K\%$ token probabilities outputted by the LLM on the sample. Note that this method requires tuning $K$, so in all our experiments we chose the best result over $K\% \in \{ 5\%, 10\%, 20\%, 30\%, 40\%, 50\%, 60\% \}$.

\subsection{LLM Watermarking Methods}

\textbf{UMD}~\citep{kirchenbauer2023watermark} splits the vocabulary into two sets (green and red) and biases the model towards the green tokens by altering the logit distribution. The hash of the previous token's ID serves as a seed for a pseudo-random number generator used to split the vocabulary into these two groups. For a ``hard'' watermark, the model is forced not to sample from the red list at all. For a ``soft'' watermark, a positive bias $\delta$ is added to the logits of the green tokens before sampling. We focus our empirical evaluation on ``soft'' watermarks as they are more suitable for LLM deployment due to their smaller impact on the quality of the generated text. \\ \\
\textbf{Unigram-Watermark}~\citep{zhao2023provable} employs a similar approach of splitting the vocabulary into two sets and biasing the model towards one of the two sets. However, the split remains consistent throughout the generation. This choice is made to provide a provable improvement against paraphrasing attacks~\citep{krishna2024paraphrasing}. \\

%% file: sections/verbatim.tex
\section{Watermarking LLMs Prevents \\Copyrighted Text Generation}\label{verbatim}

In this section, we study the effect of LLM watermarking techniques on verbatim memorization. We discuss the their implications for preventing copyrighted text generation. \\

\topic{Datasets} We consider $4$ versions of the WikiMIA benchmark~\citep{shi2023detecting} with 32, 64, 128, and 256 words in each sample and only consider the samples that were very likely part of the training set of all the models we consider (labeled as $1$ in~\citet{shi2023detecting}). We consider these subsets as a proxy for text that was used in the training set, and the model may be prone to verbatim memorization. From now on, we refer to this subset as the ``training samples'' or ``training texts''. Similarly, we consider BookMIA dataset~\citep{shi2023detecting}, which contains samples from copyrighted books. \\

\topic{Metric} We measure the relative increase in perplexity on the generation of training samples by the watermarked model compared to the original model. We report the increase in both the minimum and average perplexity over the training samples. Note that a large increase in perplexity corresponds to a large decrease in the probability of generating that specific sample, as shown later in this section. When computing the perplexity, we prompt the model with an empty string, the first 10, and the first 20 tokens of the targeted training sample, respectively. In the BookMIA dataset, we designate the initial 100 or 256 tokens as the prompt. This is because each BookMIA sample contains 512 words, which is larger than the sample size in WikiMIA. \\

\topic{Models} We conduct our empirical evaluation on $5$ recent LLMs: Llama-30B \citep{touvron2023llama}, GPT-NeoX-20B \citep{black2022gpt}, Llama-13B \citep{touvron2023llama}, Pythia-2.8B \citep{biderman2023pythia} and OPT-2.7B \citep{zhang2022opt}.

\subsection{Empirical Evaluation}

\begin{table}
  \centering
  \begin{tabular}{c|c|cc|cc}
    \toprule
    & & \multicolumn{2}{c|}{Llama-30B} & \multicolumn{2}{c}{Llama-13B} \\
    \midrule
      & P. & Min.  & Avg. & Min. & Avg. \\
    \midrule

   \multirow{3}{*}{UMD} & 0 & $ 3.3  $ & $ 31.2  $ & $ 4.9  $ & $ 34.3  $ \\
    & 10 & $ 2.8  $ & $ 28.7  $ & $ 3.5  $ & $ 31.9  $ \\

      & 20 & $ 2.4  $ & $ 30.1  $ & $ 3.5  $ & $ 33.4  $ \\

    \midrule

   \multirow{3}{*}{Unigram}       & 0 & $ 4.1  $ & $ 34.1  $ & $ 5.0  $ & $ 36.6  $  \\
    & 10 & $ 3.0  $ & $ 31.7  $ & $ 4.0  $ & $ 34.3  $ \\

          & 20 & $ 2.4  $ & $ 31.5  $ & $ 3.4  $ & $ 34.0  $  \\

    \bottomrule
  \end{tabular}
    \caption{Measuring the reduction in verbatim memorization of training texts on WikiMIA-32. We report the relative increase in both the minimum and average perplexity between the watermarked and unwatermarked models, where larger values correspond to less memorization. Note that ``P.'' stands for ``prompt length''.}
      \label{verbatim-32}

\end{table}

\begin{figure}[t]
    \centering
    \includegraphics[width=0.37\textwidth]{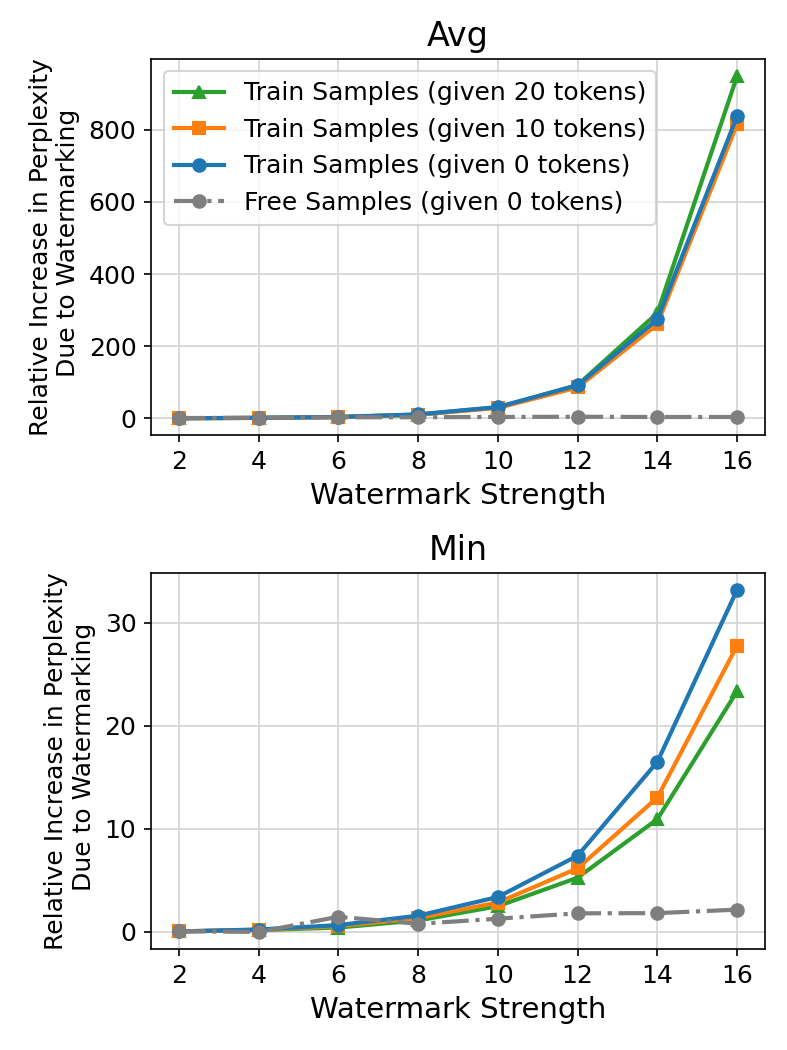}
    \caption{We study how the watermark strength (under the UMD scheme) affects the average and the minimum perplexity of training samples from WikiMIA-32, as well as the quality of generated text.}
    \label{fig:verbatim_vary_strength}
\end{figure}

In Table~\ref{verbatim-32}, we show the increase in perplexity on the training samples when the model is watermarked relative to the unwatermarked model. We observe that for Llama-30B, Unigram-Watermark induces a relative increase of $4.1$ in the minimum and $34.1$ in the average perplexity. Note that a relative increase of $4.1$ in perplexity for a sample makes it more than $4.3 \times 10^{22}$ times less likely to be generated. This is based on a sample with only $32$ tokens, which is likely a lower bound since the number of tokens is typically larger than the number of words. We observe consistent results over several models and prompt lengths. For all experiments, unless otherwise specified, we use a fixed strength parameter $\delta = 10$ for watermark methods and a fixed percentage of $50\%$ green tokens. All the results are averaged over $5$ runs with different seeds for the watermark methods. We include additional results on WikiMIA-64, WikiMIA-128 and WikiMIA-256 in Tables~7, 8 and 9, respectively, in the appendix. We observe that our findings are consistent across models and splits of WikiMIA. Finally, we include the complete version of Table~\ref{verbatim-32} in the appendix (Table~6), which shows results for additional models and random logit perturbations with the same strength as the watermarking methods. Overall, the additional results are consistent with our previous findings. \\

\begin{table}
    % \footnotesize
  \centering
  \begin{tabular}{c|c|cc|cc}
    \toprule
    & & \multicolumn{2}{c|}{Llama-30B} & \multicolumn{2}{c}{Llama-13B} \\
    \midrule
      & P. & Min.  & Avg. & Min. & Avg. \\
    \midrule

   \multirow{3}{*}{UMD}       & 0 & $ 1.5 $ & $ 33.7 $ & $ 2.4 $ & $ 41.2 $  \\
    & 10 & $ 1.5 $ & $ 33.6 $  & $ 2.3 $ & $ 41.0 $ \\

          & 20 & $ 1.4 $ & $ 33.5 $ & $ 2.3 $ & $ 40.8 $  \\
          & 100 & $ 1.3 $ & $ 32.9 $ & $ 1.9 $ & $ 40.3 $  \\

    \midrule

   \multirow{3}{*}{Unigram}   & 0 & $ 1.6  $ & $ 36.4  $ & $ 2.4 $ & $ 44.5 $  \\
    & 10 & $ 1.6 $ & $ 36.3 $ & $ 2.4 $ & $ 44.3 $  \\
      & 20 & $ 1.5 $ & $ 36.1 $ & $ 2.3 $ & $ 44.2 $ \\
      & 100 & $ 1.4 $ & $ 35.5 $ & $ 1.8 $ & $ 43.6 $ \\
    \bottomrule
  \end{tabular}
    \caption{Measuring the reduction in verbatim memorization of training texts on BookMIA. We report the relative increase in both the minimum and average perplexity between the watermarked and unwatermarked models, where larger values correspond to less memorization. Note that ``P.'' stands for ``prompt length''.}
    \label{verbatim-bookmia}

\end{table}

In Figure~\ref{fig:verbatim_vary_strength}, as well as Figure~5 from the appendix, we study the influence of the strength of the watermark $\delta$ on the relative increase in both the minimum and average perplexity on the WikiMIA-32 training samples. In this experiment, we also consider a baseline of generating text freely to study the impact of watermarks on the quality of text relative to the impact on training samples' generation (here, perplexity is computed by an unwatermarked model). All the results are averaged over $5$ runs with different seeds for the watermark methods. In the case of free generation, we generate $100$ samples for $5$ different watermarking seeds and average the results. The length of the generated samples is up to $42$ tokens, which is approximately $32$ words in the benchmark (on a token-to-word ratio of $4:3$). \textbf{The results show an exponential increase in the perplexity of the training samples with the increase in watermark strength, while the generation quality is affected at a slower rate.} This suggests that even if there is a trade-off between protecting the generation of text memorized verbatim and generating high-quality text, \textbf{finding a suitable watermark strength for each particular application is possible}. Examples of generated samples at varying watermark strengths are provided in the appendix. \\

\begin{figure}[ht!]
    \centering
    \includegraphics[width=0.40\textwidth]{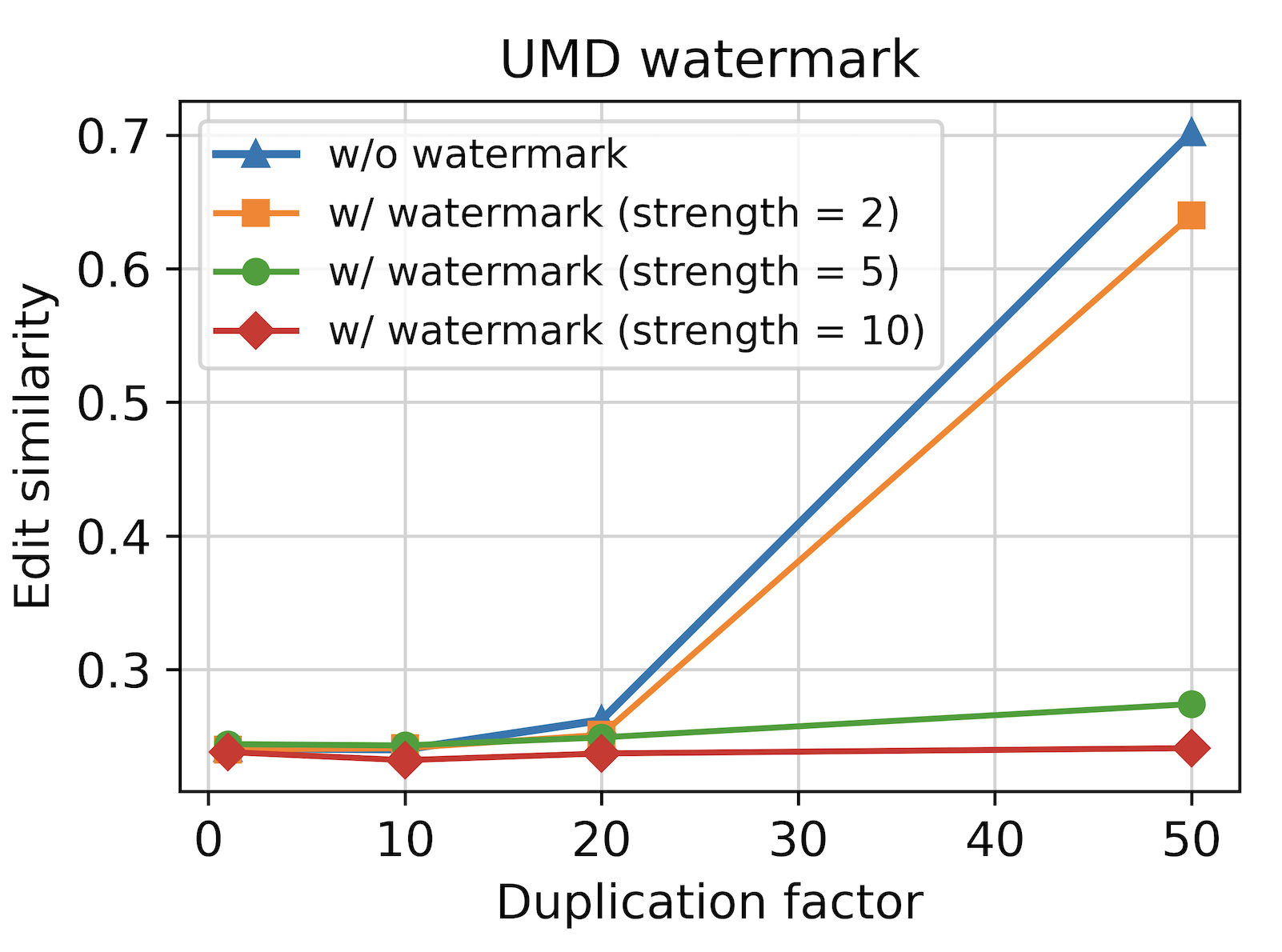}
    \caption{Edit similarity between the generated completion and the ground truth when considering different watermark strengths and memorization levels.}
    \label{fig:approximate_mem_main}
\end{figure}

\topic{Approximate Memorization} Informally, we consider a training sample approximately memorized by a model if, given its prefix, it is possible to generate a completion that is similar enough to the ground truth completion. In our experiments, we use models fine-tuned on a subset of BookMIA (details provided in the appendix) and we consider Normalized Edit Similarity (referred to as edit similarity from now on) and BLEU score as similarity measures, as in~\cite{ippolito2023preventing}. Note that we consider both word-level and token-level variants for the BLEU score. The range for each metric is between 0 and 1, where values close to 1 represent similar texts. In all experiments, since all the samples are 512 words long, we consider the first 256 words as the prefix and the last 256 words as the ground truth completion. We present the results for edit similarity with the UMD watermark in Figure~\ref{fig:approximate_mem_main}, and the complete results—using all metrics and including the Unigram watermark—in Figure~7, averaged over 20 runs with different random seeds. Note that the duplication factor (shown on x-axis) represents the number of times the target copyrighted text is duplicated. We observe that for high levels of memorization, a strong watermark significantly reduces the similarity between the generated completion and the ground truth (copyrighted) one. 

\topic{Takeaways} Watermarking significantly increases the perplexity of generating training texts, reducing verbatim memorization likelihood. This is achieved with only a moderate impact on the overall quality of generated text. This suggests that watermark strength can be effectively tailored to balance verbatim memorization and text quality for specific applications. Finally, we believe that our findings on WikiMIA---which does not necessarily contain copyrighted data---directly extend to the generation of copyrighted text verbatim, as this constitutes a form of verbatim memorization of the training data. To confirm, we run similar experiments on a dataset containing copyrighted data (BookMIA) and include the results in the Table~\ref{verbatim-bookmia}. Additionally, we consider finetuning Llama-7B~\cite{touvron2023llama} on BookMIA while controlling memorization by duplicating training samples. Detailed information about this experiment is provided in the appendix.

%% file: sections/mia.tex
\section{Impact of Watermarking on \\ Pretraining Data Detection} \label{mia}

\topic{Datasets} We revisit the WikiMIA benchmark as discussed in the previous section. We consider the full datasets, rather than the subset of samples that were part of the training for models we study. Additionally, we consider the BookMIA benchmark, which contains copyrighted texts. \\

\topic{Metrics} We follow the prior work~\citep{shi2023detecting, duarte2024cop} and report the AUC and AUC drop to study the detection performance of the MIAs. Note that this metric has the advantage of not having to tune the threshold for the detection classifier. \\

\topic{Models} We conduct experiments on the same LLMs as in the previous section. Additionally, for the Smaller Ref method that requires a smaller reference model along with the target LLM, we consider Llama-7B, Neo-125M, Pythia-70M, and OPT-350M as references.

\subsection{Empirical Evaluation}

\begin{figure}
    \centering
    \includegraphics[width=0.40\textwidth]{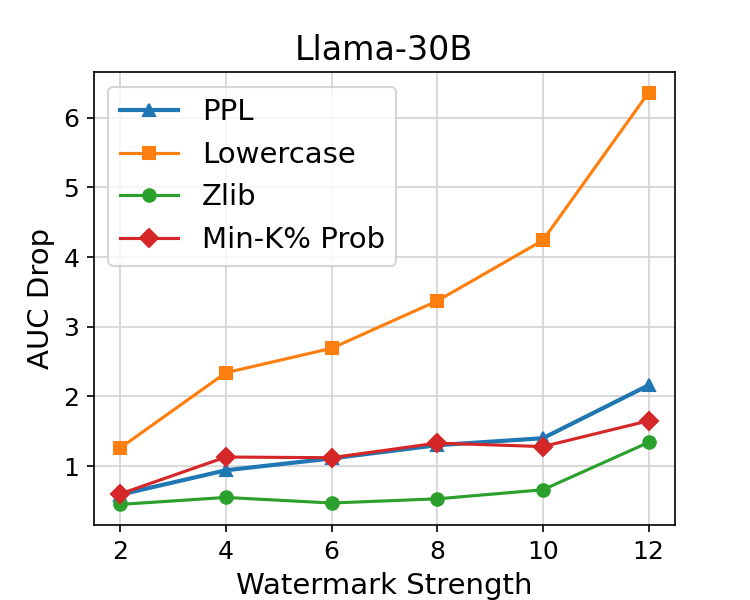}
    \caption{AUC drop due to watermarking for each MIA when varying the strength of the watermark.}
    \label{fig:mia_vary_strength}
\end{figure} 

In Table~\ref{mia-256}, we show the AUC for the unwatermarked and watermarked models using the UMD scheme, as well as the drop between the two. We observe that watermarking reduces the AUC (drop shown in bold in the table) by up to $14.2\%$ across $4$ detection methods and $5$ LLMs. All the experiments on watermarked models are run with $5$ different seeds and we report the mean and standard deviation of the results. We also report the AUC drop, which is computed by the difference between the AUC for the unwatermarked model and the mean AUC over the $5$ runs for the watermarked model. Additionally, while the experiments from Table~\ref{mia-256} are conducted on WikiMIA-256, we observe similar trends for WikiMIA-32, WikiMIA-64, and WikiMIA-128 in the appendix. We also study the impact of the watermark's strength on the AUC drop for Llama-30B in Figure~\ref{fig:mia_vary_strength} and for the other models in Figure~6 from the appendix. Note that we considered WikiMIA-256 for these experiments. We observe that higher watermark strengths generally induce larger AUC drops. \\

\begin{table}
    % \footnotesize
  \centering
  \begin{tabular}{c|cc}
    \toprule
     & Llama-30B & Llama-13B  \\
     
    \midrule
    
    \multirow{3}{*}{PPL}    & $ 85.4\%$ & $ 68.2\%$  \\
       
   & $  84.7\pm  1.4\%$ & $  67.6\pm 2.5 \%$    \\
   
       & $\textbf{ 0.7\%}$ & $\textbf{ 0.6\%}$   \\ 
       
    \midrule
    \multirow{3}{*}{Lowercase}  & $ 87.9\%$ & $ 77.6\%$  \\
    
                & $  80.9\pm 3.1\%$ & $  67.2\pm 4.0 \%$    \\
   
                & $\textbf{ 7.0\%}$ & $\textbf{ 10.4\%}$    \\
       
    \midrule
    
   \multirow{3}{*}{Zlib}  & $ 82.5\%$ & $ 62.5\%$  \\
    
   & $  77.8\pm  1.2\%$ & $ 57.1 \pm 2.0 \%$    \\
   
       & $\textbf{ 4.7\%}$ & $\textbf{ 5.4\%}$ \\
       
    \midrule
    
   \multirow{3}{*}{Min-K\% Prob} & $ 85.1\%$ & $ 70.2\%$  \\
    
    & $  85.0\pm  1.0\%$ & $ 68.5 \pm 0.1 \%$    \\
   
       & $\textbf{ 0.1\%}$ & $\textbf{ 1.7\%}$     \\

    \bottomrule
  \end{tabular}
    \caption{AUC of each MIA for the unwatermarked (\textit{top} of each cell), watermarked models (\textit{middle} of each cell), and the drop between the two (\textit{bottom} of each cell) on BookMIA using UMD scheme.}
  \label{mia-bookmia}
\end{table}

\begin{table*}

  \centering
  \begin{tabular}{c|ccccc}
    \toprule
     & Llama-30B & NeoX-20B & Llama-13B & Pythia-2.8B & OPT-2.7B \\
    \midrule
       & $72.0\%$ & $71.3\%$ & $71.2\%$ & $67.8\%$ & $60.5\%$   \\
   PPL & $70.6 \pm 1.9\%$ & $64.7 \pm 2.3\%$ & $70.0 \pm 2.6\%$ & $64.4 \pm 1.9\%$ & $54.9 \pm 2.2\%$   \\
       & $\textbf{1.4\%}$ & $\textbf{6.6\%}$ & $\textbf{1.2\%}$  & $\textbf{3.4\%}$  & $\textbf{5.6\%}$    \\
    \midrule
             & $68.1\%$ & $68.2\%$ & $65.5\%$ & $62.9\%$ & $58.9\%$   \\
   Lowercase & $63.8 \pm 4.5\%$ & $55.4 \pm 5.5\%$ & $61.6 \pm 3.8\%$ & $58.7 \pm 3.2\%$ & $49.7 \pm 2.9\%$   \\
              & $\textbf{4.3\%}$  & $\textbf{14.2\%}$  & $\textbf{3.9\%}$  & $\textbf{4.2\%}$  & $\textbf{9.2\%}$    \\
    \midrule
        & $72.7\%$ & $73.2\%$ & $73.1\%$ & $69.2\%$ & $62.7\%$   \\
   Zlib & $72.0 \pm 1.6\%$ & $66.6 \pm 2.0\%$ & $71.6 \pm 2.3\%$ & $66.1 \pm 1.2\%$ & $58.1 \pm 1.8\%$   \\
        & $\textbf{0.7\%}$ & $\textbf{6.6\%}$  & $\textbf{1.5\%}$  & $\textbf{3.1\%}$  & $\textbf{4.6\%}$     \\
    \midrule
             & $71.8\%$ & $78.0\%$ & $72.9\%$ & $71.0\%$ & $65.5\%$   \\
Min-K\% Prob & $70.5 \pm 1.8\%$ & $76.2 \pm 2.1\%$ & $70.4 \pm 3.2\%$ & $69.5 \pm 1.6\%$ & $63.1 \pm 3.4\%$   \\
             & $\textbf{1.3\%}$ & $\textbf{1.8\%}$ & $\textbf{2.5\%}$ & $\textbf{1.5\%}$ & $\textbf{2.4\%}$    \\
   
    \bottomrule
  \end{tabular}
    \caption{AUC of each MIA for the unwatermarked (\textit{top} of each cell), watermarked models (\textit{middle} of each cell), and the drop between the two (\textit{bottom} of each cell) on WikiMIA-256 using UMD scheme.}
  \label{mia-256}
\end{table*}

In addition to the $4$ detection methods, we also consider Smaller Ref attack, which we include in Table~13 of the appendix. We consider different variations, including an unwatermarked reference model and a watermarked one with a similar strength but a different seed or with both strength and seed changed in comparison to the watermarked target model. The baseline is an unwatermarked model with an unwatermarked reference model. We observe the AUC drops in all scenarios (up to $16.4\%$), which is consistent with our previous findings. \\ \\
We also experiment with several percentages of green tokens for a fixed watermark strength of $\delta = 10$. We show the results in Table~14 of the appendix. We observe that for all models, in at least $80\%$ of the cases all of the attacks' AUCs are negatively affected (positive drop value), suggesting that, in general, \textbf{finding a watermarking scheme that reduces the success rates of the current MIAs is not a difficult task}. Note that the experiments are run on WikiMIA for UMD scheme and the results are averaged over $5$ watermark seeds. \\ \\
\topic{Takeaways} Watermarking can significantly reduce the success of membership inference attacks (MIAs), with AUC drops up to $16.4\%$. By varying the percentage of green tokens as well as the watermark's strength, we observe that watermarking schemes can be easily tuned to negatively impact the detection success rates of MIAs. Finally, we conduct experiments on the BookMIA dataset and observe results consistent with our previous findings. These results are included in Table~\ref{mia-bookmia}.

%% file: sections/method.tex
\section{Improving Detection Performance  with Adaptive Min-K\% Prob}

\label{adaptive}

This section demonstrates how an informed, adaptive attacker can improve the success rate of a recent MIA, Min-K\% Prob. Our main idea is that an attacker with knowledge of the watermarking technique (including green-red token lists and watermark's strength $\delta$) can readjust token probabilities. This is possible even without additional information about the logit distribution, relying solely on the probability of each token from the target sample given the preceding tokens. Our approach relies on two key assumptions. First, knowledge of the watermarking scheme, which aligns with assumptions made in prior work on public watermark detection~\citep{kirchenbauer2023watermark}. Second, access to the probability of each token in a sample, given the previous tokens—an assumption also made by the Min-K\% Prob method \cite{shi2023detecting}. \\

\topic{Threat model} (1) The attacker's goal is to infer whether specific samples are part of the training set or not. In our setting, the attacker is not malicious, as the goal is to detect copyright violations. (2) Regarding the attacker's knowledge, we assume the attacker knows the watermarking method and its parameters (green and red lists, and the watermark strength), which aligns with the assumption made in prior work by~\citet{kirchenbauer2023watermark} for public watermark detection. (3) As for the attacker's capabilities, we assume they can access the probabilities for each token in the given samples, similar to what a copyright auditor may have access to. This also mirrors the assumption made by~\citet{shi2023detecting} in the context of training data detection. \\

Our method described in Algorithm~\ref{alg:1} is based on the observation that if the denominator of softmax function (i.e., $\sum_i e^{z_i}$, where $z_i$ is the logit for the $i$-th vocabulary) does not vary significantly when generating samples with the watermarked model (and similarly for the unwatermarked model), then we can readjust the probabilities of the green tokens by ``removing'' the bias $\delta$. More precisely, assuming the approximation for the denominator of softmax is good, then the probability for each token $t_i$ in an unwatermarked model will be around $\frac{e^{L_i}}{c}$, where $L_i$ is the logit corresponding to the token $t_i$ and $c$ is a constant. However, for a watermarked model, if the token $t_i$ is green, then the probability would be approximated by $\frac{e^{L_i+\delta}}{d}$, where $d$ is again a constant, while in the case $t_i$ is red the probability will be around $\frac{e^{L_i}}{d}$. To compensate for the bias introduced by watermarking, we divide the probability of green tokens by $e^{\delta}$ and this way we end up with probabilities that are just a scaled (by $\frac{c}{d}$) version of the probabilities from the unwatermarked model. The scaling factor will not affect the orders between the samples when computing the average of the minimum K\% log-probabilities as long as the tested sentences are approximately the same length, which is an assumption made by~\citet{shi2023detecting} as well. \\ 

Despite the strong assumption we assumed regarding the approximation of the denominator, empirical results show that our method effectively improves the success rate of Min-K\% under watermarking. We show results for two LLMs in Table~\ref{ours} and include the complete results for 5 LLMs in Table~17 from the appendix. We observe that our method improves over the baseline in $95\%$ of the cases, and the increase is as high as $4.8\%$ (averaged over 5 runs). \\

Finally, we also consider adaptive versions of the Lowercase and Zlib methods. Our findings show that these adaptive methods outperform the baselines in at least 80\% of cases. Detailed results are provided in the appendix.\\

\topic{Takeaways} We demonstrate that an adaptive attacker can leverage the knowledge of a watermarking scheme to increase the success rates of recent MIAs.

\begin{algorithm*}[ht!]
% \small
\caption{Adaptive Min-K\% Prob}\label{alg:1}
\DontPrintSemicolon
\ResetInOut{Output} 
\ResetInOut{Require} 
\Require{
Tokenized target sample $t = t_1 \oplus t_2 \oplus ... \oplus t_n$, access to the probability of the target (watermarked) LLM $f$ to generate $t_i$ given the $i-1$ previous tokens and $t_0$ (empty string) $f(t_i | t_0 \oplus t_1 \oplus ... \oplus t_{i-1})$ (similar assumption as Min-K\% Prob algorithm), $K$, we assume we know the watermarking scheme (e.g., for public watermark detection purposes), i.e. we know the green and red lists as well as $\delta$.
}
\Output{
Adjusted average of the minimum $K\%$ token probabilities when generating $t_1 \oplus t_2 \oplus ... \oplus t_n$
}
adj\_prob $\gets \{ \}$\hfill\Comment{\textcolor{gray}{The set of adjusted probabilities}}
% \If{algorithm\_type == oracle}{
\For{$i  \in 1, 2, \dots, n$}{
    $p_f(t_i) \gets f(t_i | t_0 \oplus t_1 \oplus ... \oplus t_{i-1})$ \hfill\Comment{\textcolor{gray}{Probability of $t_i$ when the model is watermarked}} 
        \eIf{$t_i$ is green}{
    adj\_prob $\gets$ adj\_prob $\cup \{ \frac{p_f(t_i)}{e^{\delta} } \}$ \hfill\Comment{\textcolor{gray}{Adjust the probability if the token is green}}
    }
    {adj\_prob $\gets$ adj\_prob $\cup \{ p_f(t_i) \}$ }
}

$k = floor(n \cdot K\%)$ \hfill\Comment{\textcolor{gray}{Find the number of token probabilities to keep}}
adj\_k\_prob $\gets min\_k$(adj\_prob)\hfill\Comment{\textcolor{gray}{Select the minimum $k$ probabilities}}
\Return mean(log(adj\_k\_prob) )\hfill\Comment{\textcolor{gray}{Return the mean of the minimum $k$ log-probabilities}}

\end{algorithm*}

\begin{table}
    % \footnotesize

  \centering
  \begin{tabular}{c|c|cc}
    \toprule
    & & Llama-30B & Llama-13B \\
    \midrule

   WikiMIA    & Not adapt. & $66.2\%$ & $64.5\%$ \\
   32 & Adapt. & $\textbf{68.5\%}$ & $\textbf{66.3\%}$ \\

    \midrule
    WikiMIA   & Not adapt. & $64.4\%$ & $62.8\%$ \\
   64 & Adapt. & $\textbf{67.3\%}$ & $\textbf{64.9\%}$  \\

    \midrule
    WikiMIA   & Not adapt. & $70.0\%$ & $68.9\%$  \\
   128 & Adapt. & $\textbf{73.1\%}$ &  $\textbf{71.0\%}$ \\

    \midrule
     WikiMIA  & Not adapt. & $70.5\%$ & $70.4\%$  \\
   256 & Adapt. & $\textbf{71.3\%}$ & $\textbf{72.4\%}$  \\

    \bottomrule
  \end{tabular}
    \caption{We show the AUC of Min-\%K Prob (referred as ``Not adapt.'') and our method (referred as ``Adapt.'') when using UMD watermarking scheme. We highlight the cases when our method improves over the baseline.}
  \label{ours}
\end{table}

%% file: sections/conclusion.tex
\section{Conclusion and Discussion}\label{conclusion}

Watermarking LLMs has unintended consequences on methods towards copyright protection. Our experiments demonstrate that while watermarking may be a promising solution to prevent copyrighted text generation, watermarking also complicates membership inference attacks that may be employed to detect copyright abuses. Watermarking can be a double-edged sword for copyright regulators since it promotes compliance during generation time, while making training time copyright violations harder to detect. We hope our work furthers the discussion around watermarking and copyright issues for LLMs.

%% file: sections/ack.tex
\section*{Acknowledgements}
Panaitescu-Liess, Che, An, Xu, Pathmanathan, Chakraborty, Zhu, and Huang are supported by DARPA Transfer from Imprecise and Abstract Models to Autonomous Technologies (TIAMAT) 80321, National Science Foundation NSF-IIS-2147276 FAI, DOD-AFOSR-Air Force Office of Scientific Research under award number FA9550-23-1-0048, Adobe, Capital One and JP Morgan faculty fellowships.

%% file: sections/appendix.tex
\appendix

\onecolumn

\section{Appendix}

\vspace{10px}

\label{appendix}

\subsection{Additional experiments on verbatim memorization on WikiMIA}

\label{appendix-verbatim}

\begin{table}[ht!]
    % \footnotesize
  \caption{Measuring the reduction in verbatim memorization of training texts on WikiMIA-32. We report the relative increase in both the minimum and average perplexity between the watermarked and unwatermarked models, where larger values correspond to less memorization. Note that ``P.'' stands for ``prompt length''.}
  \label{verbatim-32-full}
  \centering
  \begin{tabular}{c|c|cc|cc|cc|cc|cc}
    \toprule
    & & \multicolumn{2}{c|}{Llama-30B} & \multicolumn{2}{c|}{NeoX-20B} & \multicolumn{2}{c|}{Llama-13B} & \multicolumn{2}{c|}{Pythia-2.8B} & \multicolumn{2}{c}{OPT-2.7B} \\
    \midrule
      & P. & Min.  & Avg. & Min. & Avg. & Min. & Avg. & Min. & Avg. & Min. & Avg. \\
    \midrule

    & 0 & $ 3.3  $ & $ 31.2  $ & $ 3.7  $ & $ 52.1  $ & $ 4.9  $ & $ 34.3  $ & $ 11.4  $ & $ 61.3  $ & $ 10.4  $ & $ 64.5  $  \\
   UMD & 10 & $ 2.8  $ & $ 28.7  $ & $ 2.2  $ & $ 52.1  $ & $ 3.5  $ & $ 31.9  $ & $ 8.8  $ & $ 63.7  $ & $ 8.3  $ & $ 67.7  $ \\

      & 20 & $ 2.4  $ & $ 30.1  $ & $ 1.8  $ & $ 66.0  $ & $ 3.5  $ & $ 33.4  $ & $ 5.0  $ & $ 74.0  $ & $ 7.0  $ & $ 84.4  $ \\

    \midrule

          & 0 & $ 4.1  $ & $ 34.1  $ & $ 4.4  $ & $ 54.1  $ & $ 5.0  $ & $ 36.6  $ & $ 14.3  $ & $ 74.5  $ & $ 11.5  $ & $ 66.1  $  \\
   Unigram & 10 & $ 3.0  $ & $ 31.7  $ & $ 2.8  $ & $ 52.5  $ & $ 4.0  $ & $ 34.3  $ & $ 11.8  $ & $ 73.6  $ & $ 9.8  $ & $ 70.2  $ \\

          & 20 & $ 2.4  $ & $ 31.5  $ & $ 2.0  $ & $ 56.4  $ & $ 3.4  $ & $ 34.0  $ & $ 6.6  $ & $ 79.1  $ & $ 5.8  $ & $ 81.4  $ \\

    \midrule

      & 0 & $ 4.0 $ & $ 34.3 $ & $ 4.9 $ & $ 51.1 $ & $ 5.5 $ & $ 34.7 $ & $ 8.0 $ & $ 62.3 $ & $ 7.4 $ & $ 60.6 $ \\
   Random & 10 & $ 2.6 $ & $ 31.4 $ & $ 3.1 $ & $ 51.4 $ & $ 3.6 $ & $ 31.7 $ & $ 5.5 $ & $ 62.9 $ & $ 6.3 $ & $ 64.0 $ \\

      & 20 & $ 2.1 $ & $ 31.8 $ & $ 1.2 $ & $ 59.3 $ & $ 2.8 $ & $ 32.9 $ & $ 4.7 $ & $ 78.6 $ & $ 3.9 $ & $ 73.7 $ \\

    \bottomrule
  \end{tabular}
\end{table}

\begin{table}[ht!]
    % \footnotesize
  \caption{Measuring the reduction in verbatim memorization of training texts on WikiMIA-64. We report the relative increase in both the minimum and average perplexity between the watermarked and unwatermarked models, where larger values correspond to less memorization. Note that ``P.'' stands for ``prompt length''.}
  \label{verbatim-64}
  \centering
  \begin{tabular}{c|c|cc|cc|cc|cc|cc}
    \toprule
    & & \multicolumn{2}{c|}{Llama-30B} & \multicolumn{2}{c|}{NeoX-20B} & \multicolumn{2}{c|}{Llama-13B} & \multicolumn{2}{c|}{Pythia-2.8B} & \multicolumn{2}{c}{OPT-2.7B} \\
    \midrule
      & P. & Min.  & Avg. & Min. & Avg. & Min. & Avg. & Min. & Avg. & Min. & Avg. \\
    \midrule

      & 0 & $ 4.9  $ & $ 27.6  $ & $ 4.2  $ & $ 42.8  $ & $ 6.7  $ & $ 30.5  $ & $ 15.2  $ & $ 50.4  $ & $ 14.9  $ & $ 51.7  $ \\
   UMD & 10 & $ 4.3  $ & $ 26.2  $ & $ 3.7  $ & $ 41.3  $ & $ 6.1  $ & $ 29.1  $ & $ 15.6  $ & $ 49.2  $ & $ 14.4  $ & $ 51.4  $  \\

      & 20 & $ 3.9  $ & $ 26.4  $ & $ 3.6  $ & $ 43.1  $ & $ 5.8  $ & $ 29.3  $ & $ 14.0  $ & $ 50.3  $ & $ 12.5  $ & $ 52.7  $  \\

    \midrule

          & 0 & $ 5.0  $ & $ 28.1  $ & $ 4.3  $ & $ 45.3  $ & $ 6.7  $ & $ 30.9  $ & $ 17.6  $ & $ 62.3  $ & $ 16.0  $ & $ 53.7  $ \\
   Unigram & 10 & $ 3.8  $ & $ 26.9  $ & $ 3.4  $ & $ 43.6  $ & $ 5.3  $ & $ 29.7  $ & $ 16.2  $ & $ 60.6  $ & $ 17.1  $ & $ 53.0  $  \\

          & 20 & $ 3.2  $ & $ 26.9  $ & $ 3.1  $ & $ 44.2  $ & $ 4.4  $ & $ 29.7  $ & $ 13.6  $ & $ 60.9  $ & $ 11.7  $ & $ 53.6  $  \\
      
    \bottomrule
  \end{tabular}
\end{table}

\begin{table}[ht!]
    % \footnotesize
  \caption{Measuring the reduction in verbatim memorization of training texts on WikiMIA-128. We report the relative increase in both the minimum and average perplexity between the watermarked and unwatermarked models, where larger values correspond to less memorization. Note that ``P.'' stands for ``prompt length''.}
  \label{verbatim-128}
  \centering
  \begin{tabular}{c|c|cc|cc|cc|cc|cc}
    \toprule
    & & \multicolumn{2}{c|}{Llama-30B} & \multicolumn{2}{c|}{NeoX-20B} & \multicolumn{2}{c|}{Llama-13B} & \multicolumn{2}{c|}{Pythia-2.8B} & \multicolumn{2}{c}{OPT-2.7B} \\
    \midrule
      & P. & Min.  & Avg. & Min. & Avg. & Min. & Avg. & Min. & Avg. & Min. & Avg. \\
    \midrule

      & 0 & $ 5.7  $ & $ 25.3  $ & $ 4.6  $ & $ 39.5  $ & $ 7.6  $ & $ 28.0  $ & $ 23.1  $ & $ 45.3  $ & $ 18.6  $ & $ 48.1  $  \\
   UMD & 10 & $ 5.3  $ & $ 24.4  $ & $ 4.3  $ & $ 38.9  $ & $ 7.2  $ & $ 27.1  $ & $ 23.6  $ & $ 44.7  $ & $ 19.1  $ & $ 47.6  $ \\

      & 20 & $ 5.2  $ & $ 24.5  $ & $ 4.3  $ & $ 39.3  $ & $ 6.8  $ & $ 27.2  $ & $ 23.0  $ & $ 44.7  $ & $ 17.5  $ & $ 47.8  $ \\

    \midrule

          & 0 & $ 4.5  $ & $ 25.6  $ & $ 5.9  $ & $ 42.9  $ & $ 6.4  $ & $ 28.2  $ & $ 17.6  $ & $ 54.9  $ & $ 19.6  $ & $ 50.0  $ \\
   Unigram & 10 & $ 3.9  $ & $ 25.0  $ & $ 5.3  $ & $ 42.0  $ & $ 5.7  $ & $ 27.6  $ & $ 15.8  $ & $ 53.6  $ & $ 18.7  $ & $ 49.9  $  \\

          & 20 & $ 3.6  $ & $ 25.2  $ & $ 5.1  $ & $ 42.1  $ & $ 5.3  $ & $ 27.7  $ & $ 15.1  $ & $ 53.6  $ & $ 18.0  $ & $ 49.9  $  \\
      
    \bottomrule
  \end{tabular}
\end{table}

\begin{table}[ht!]
    % \footnotesize
  \caption{Measuring the reduction in verbatim memorization of training texts on WikiMIA-256. We report the relative increase in both the minimum and average perplexity between the watermarked and unwatermarked models, where larger values correspond to less memorization. Note that ``P.'' stands for ``prompt length''.}
  \label{verbatim-256}
  \centering
  \begin{tabular}{c|c|cc|cc|cc|cc|cc}
    \toprule
     & & \multicolumn{2}{c|}{Llama-30B} & \multicolumn{2}{c|}{NeoX-20B} & \multicolumn{2}{c|}{Llama-13B} & \multicolumn{2}{c|}{Pythia-2.8B} & \multicolumn{2}{c}{OPT-2.7B} \\
    \midrule
      & P. & Min.  & Avg. & Min. & Avg. & Min. & Avg. & Min. & Avg. & Min. & Avg. \\
    \midrule

     & 0 & $ 7.5  $ & $ 23.9  $ & $ 15.4  $ & $ 37.8  $ & $ 13.0  $ & $ 26.3  $ & $ 31.2  $ & $ 45.4  $ & $ 27.3  $ & $ 46.3  $  \\
   UMD & 10 & $7.3  $ & $23.4  $ & $15.5  $ & $37.6  $ & $12.5  $ & $25.8  $ & $30.9  $ & $45.2  $ & $27.5  $ & $46.0  $ \\

      & 20 & $7.2  $ & $23.5  $ & $16.1  $ & $37.6  $ & $12.6  $ & $25.9  $ & $30.4  $ & $45.0  $ & $27.6  $ & $46.2  $ \\

    \midrule

         & 0  & $7.4  $ & $24.4  $ & $21.0  $ & $42.4  $ & $13.9  $ & $26.8  $ & $36.9  $ & $54.3  $ & $28.8  $ & $46.3  $ \\
   Unigram & 10 & $7.1  $ & $24.1  $ & $21.2  $ & $41.9  $ & $13.7  $ & $26.5  $ & $35.4  $ & $53.7  $ & $28.2  $ & $46.0  $ \\

         & 20  & $6.7  $ & $24.2  $ & $21.5  $ & $41.8  $ & $13.7  $ & $26.5  $ & $34.6  $ & $53.4  $ & $29.3  $ & $45.9  $ \\

    \bottomrule
  \end{tabular}
\end{table}

\begin{figure}[ht!]
    \centering
    \includegraphics[width=0.8\textwidth]{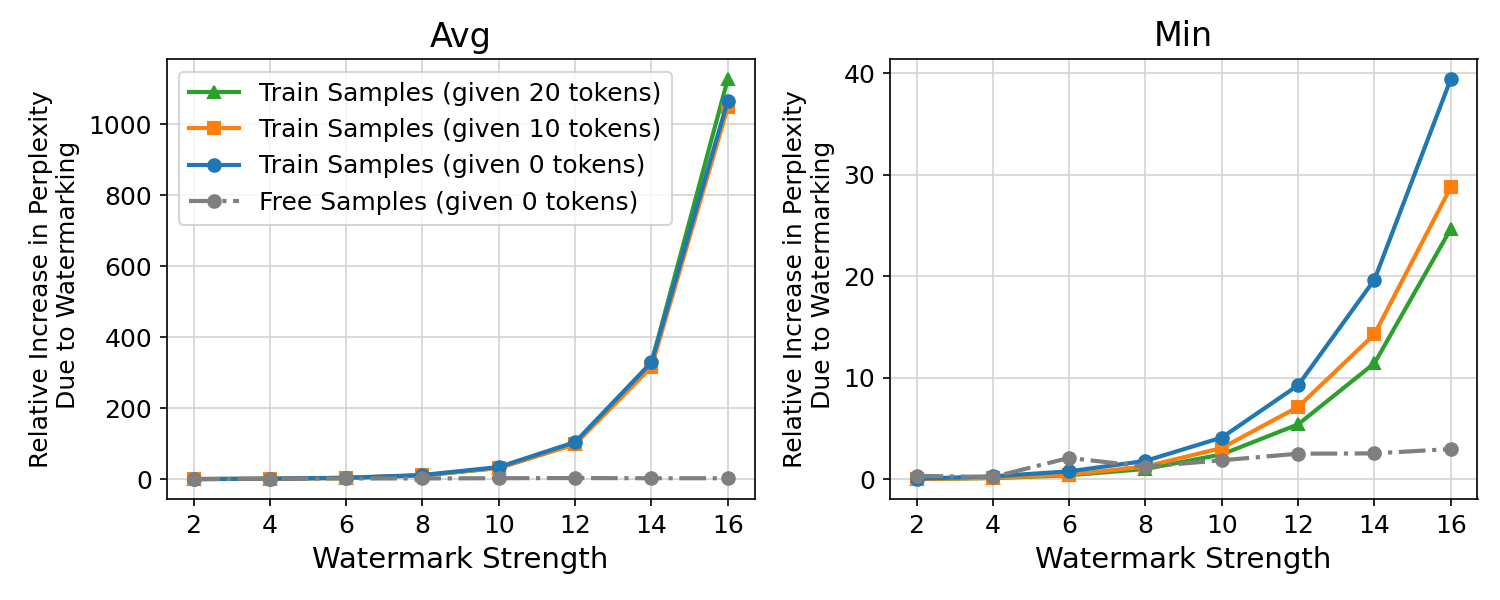}
    \caption{We study how the watermark strength (under the Unigram scheme) affects the average and the minimum perplexity of training samples from WikiMIA-32, as well as the quality of generated text.}
    \label{fig:verbatim_vary_strength2}
\end{figure}

\vspace{100px}

\subsection{Additional experiments on pretraining data detection on WikiMIA}

\label{appendix-mia}

\begin{table}[ht!]
    % \footnotesize
  \caption{AUC of each MIA for the unwatermarked (\textit{top} of each cell), watermarked models (\textit{middle} of each cell) and the drop between the two (\textit{bottom} of each cell) on WikiMIA-128 using UMD scheme.}
  \label{sample-table-1}
  \centering
  \begin{tabular}{c|ccccc}
    \toprule
     & Llama-30B & NeoX-20B & Llama-13B & Pythia-2.8B & OPT-2.7B \\
    \midrule
       & $70.3\% $ & $70.6\% $ & $67.7\% $ & $62.8\% $ & $60.0\% $   \\
   PPL & $66.3 \pm 2.2\% $ & $63.6 \pm 2.4\% $ & $63.4 \pm 2.6\% $ & $61.4 \pm 2.3\% $ & $55.1 \pm 1.6\% $   \\
       & $\textbf{4.0\%}$ & $\textbf{7.0\%}$  & $\textbf{4.3\%}$  & $\textbf{1.4\%}$   & $\textbf{4.9\%}$   \\
    \midrule
             & $59.1\% $ & $68.0\% $ & $60.6\% $ & $59.4\% $ & $57.1\% $   \\
   Lowercase & $55.9 \pm 2.9\% $ & $58.2 \pm 3.4\% $ & $55.1 \pm 3.0\% $ & $55.7 \pm 1.6\% $ & $49.2 \pm 4.5\% $   \\
             & $\textbf{3.2\%}$ & $\textbf{9.2\%}$ & $\textbf{5.5\%}$ & $\textbf{3.7\%}$ & $\textbf{7.9\%}$   \\
    \midrule
        & $71.8\% $ & $72.3\% $ & $69.6\% $ & $64.9\% $ & $62.3\% $   \\
   Zlib & $68.6 \pm 2.3\% $ & $66.3 \pm 2.1\% $ & $65.8 \pm 2.7\% $ & $63.9 \pm 1.9\% $ & $58.9 \pm 1.3\% $   \\
        & $\textbf{3.2\%}$ & $\textbf{6.0\%}$ & $\textbf{3.8\%}$ & $\textbf{1.0\%}$ & $\textbf{3.4\%}$   \\
    \midrule
             & $73.8\% $ & $76.4\% $ & $71.5\% $ & $66.8\% $ & $64.3\% $   \\
Min-K\% Prob & $70.0 \pm 1.5\% $ & $72.8 \pm 2.3\% $ & $68.9 \pm 2.2\% $ & $64.8 \pm 1.4\% $ & $59.2 \pm 2.4\% $   \\
             & $\textbf{3.8\%}$ & $\textbf{3.6\%}$ & $\textbf{2.6\%}$ & $\textbf{2.0\%}$ & $\textbf{5.1\%}$   \\
   
    \bottomrule
  \end{tabular}
\end{table}

\begin{table}[ht!]
% \footnotesize
\caption{AUC of each MIA for the unwatermarked (\textit{top} of each cell), watermarked models (\textit{middle} of each cell) and the drop between the two (\textit{bottom} of each cell) on WikiMIA-64 using UMD scheme.}  \label{sample-table-2}
  \centering
  \begin{tabular}{c|ccccc}
    \toprule
     & Llama-30B & NeoX-20B & Llama-13B & Pythia-2.8B & OPT-2.7B \\
    \midrule
       & $66.1\% $ & $66.6\% $ & $63.6\% $ & $58.4\% $ & $55.1\% $   \\
   PPL & $60.7 \pm 3.4\% $ & $60.1 \pm 3.2\% $ & $58.0 \pm 3.7\% $ & $58.7 \pm 1.7\% $ & $52.2 \pm 2.1\% $   \\
       & $\textbf{5.4\%}$ & $\textbf{6.5\%}$ & $\textbf{5.6\%}$& $\textbf{-0.3\%}$ & $\textbf{2.9\%}$    \\
    \midrule
             & $61.8\% $ & $66.4\% $ & $62.0\% $ & $57.7\% $ & $56.6\% $   \\
   Lowercase & $54.8 \pm 1.7\% $ & $56.8 \pm 3.8\% $ & $53.8 \pm 1.1\% $ & $54.5 \pm 1.0\% $ & $51.4 \pm 3.1\% $   \\
             & $\textbf{7.0\%}$ & $\textbf{9.6\%}$ & $\textbf{8.2\%}$ & $\textbf{3.2\%}$ & $\textbf{5.2\%}$    \\
    \midrule
        & $67.4\% $ & $68.1\% $ & $65.3\% $ & $60.5\% $ & $57.7\% $   \\
   Zlib & $62.4 \pm 3.3\% $ & $62.0 \pm 2.6\% $ & $59.9 \pm 3.6\% $ & $60.9 \pm 1.8\% $ & $55.5 \pm 1.5\% $   \\
        & $\textbf{5.0\%}$& $\textbf{6.1\%}$& $\textbf{4.9\%}$& $\textbf{5.4\%}$ & $\textbf{2.2\%}$    \\
    \midrule
             & $68.4\% $ & $72.8\% $ & $65.9\% $ & $61.2\% $ & $58.0\% $   \\
Min-K\% Prob & $64.4 \pm 2.9\% $ & $67.7 \pm 3.3\% $ & $62.8 \pm 3.4\% $ & $59.8 \pm 0.7\% $ & $55.3 \pm 2.3\% $   \\
             & $\textbf{4.0\%}$& $\textbf{5.1\%}$ & $\textbf{3.1\%}$ & $\textbf{1.4\%}$ &  $\textbf{2.7\%}$  \\
   
    \bottomrule
  \end{tabular}
\end{table}

\begin{table}[ht!]
% \footnotesize
\caption{AUC of each MIA for the unwatermarked (\textit{top} of each cell), watermarked models (\textit{middle} of each cell) and the drop between the two (\textit{bottom} of each cell) on WikiMIA-32 using UMD scheme.}  \label{sample-table}
  \centering
  \begin{tabular}{c|ccccc}
    \toprule
     & Llama-30B & NeoX-20B & Llama-13B & Pythia-2.8B & OPT-2.7B \\
    \midrule
       & $69.4\% $ & $69.0\% $ & $67.5\% $ & $61.3\% $ & $58.2\% $   \\
   PPL & $63.6 \pm 5.2\% $ & $62.7 \pm 3.5\% $ & $61.4 \pm 5.7\% $ & $60.8 \pm 2.3\% $ & $55.2 \pm 2.1\% $   \\
       & $\textbf{5.5\%}$ & $\textbf{6.3\%}$& $\textbf{6.1\%}$ & $\textbf{0.5\%}$ & $\textbf{3.0\%}$   \\
    \midrule
             & $64.1\% $ & $68.2\% $ & $63.9\% $ & $60.9\% $ & $59.2\% $   \\
   Lowercase & $54.9 \pm 1.8\% $ & $59.4 \pm 4.8\% $ & $54.2 \pm 1.8\% $ & $55.5 \pm 1.6\% $ & $52.1 \pm 3.9\% $   \\
             & $\textbf{9.2\%}$& $\textbf{8.8\%}$ & $\textbf{9.7\%}$ & $\textbf{0.6\%}$ & $\textbf{2.8\%}$   \\
    \midrule
        & $69.8\% $ & $69.2\% $ & $67.8\% $ & $62.1\%$ & $59.4\% $   \\
   Zlib & $64.4 \pm 4.7\% $ & $63.2 \pm 2.8\% $ & $62.3 \pm 5.1\% $ & $61.5 \pm 1.9\% $ & $56.6 \pm 1.6\% $   \\
        & \textbf{5.4\%} & \textbf{6.0\%} & \textbf{5.5\%} & \textbf{0.6\%} & \textbf{2.8\%}    \\
    \midrule
             & $70.1\% $ & $72.1\% $ & $67.9\% $ & $61.8\% $ & $59.2\% $   \\
Min-K\% Prob & $66.2 \pm 4.2\% $ & $67.1 \pm 4.2\% $ & $64.5 \pm 4.1\% $ & $61.0 \pm 1.5\% $ & $55.8 \pm 2.3\% $   \\
             & \textbf{3.9\%} & \textbf{5.0\%} & \textbf{3.4\%} & \textbf{0.8\%} &  \textbf{3.4\%}  \\
   
    \bottomrule
  \end{tabular}
\end{table}

\begin{table}[ht!]
% \footnotesize
  \caption{Results for Smaller Ref attack on WikiMIA-256. The first two rows represent the pair of target and smaller reference model, ``No model w.'' row represents the baseline AUC of a unwatermarked target LLM and unwatermarked reference model, the other three ``double rows'' correspond to different variations of the reference model and each cell contains the AUC followed by the AUC drop in comparison to the baseline.}
  \label{mia-smaller-ref}
  \centering
  \begin{tabular}{c|ccccc}
    \toprule
    & Llama-30B & NeoX-20B & Llama-13B & Pythia-2.8B & OPT-2.7B \\
    \midrule
    & Llama-7B & Neo-125M & Llama-7B & Pythia-70M & OPT-350M \\
    \midrule
     No model w. & $74.7\%$ & $70.2\% $ & $70.5\% $ & $63.6\% $ & $64.4\% $ \\
    \midrule
     Ref. not w. & $69.7 \pm 3.3\% $ & $61.0 \pm 1.8\% $ & $66.3 \pm 4.6\% $ & $61.6 \pm 2.0\% $ & $53.2 \pm 3.4\%$ \\
      & $\textbf{5.0\%} $ & $\textbf{9.2\%} $ & $ \textbf{4.2\%}$ & $\textbf{2.0\%} $ & $\textbf{11.2\%} $ \\
    \midrule
     Ref. diff. seed & $61.7 \pm 4.4\% $ & $55.5 \pm 3.4\% $ & $54.1 \pm 4.4\% $ & $58.3 \pm 2.4\% $ & $51.3 \pm 4.3\% $ \\
      & $ \textbf{13.0\%}$ & $ \textbf{15.0\%}$ & $\textbf{16.4\%} $ & $\textbf{5.3\%} $ & $\textbf{13.1\%} $ \\
    \midrule
     Ref. diff. str. & $73.7 \pm 2.6\% $ & $61.0 \pm 3.2\% $ & $68.8 \pm 4.8\% $ & $62.5 \pm 1.2\% $ & $57.3 \pm 3.6\% $ \\
      & $ \textbf{1.0\%}$ & $ \textbf{9.2\%} $ & $ \textbf{1.7\%} $ & $ \textbf{1.1\%}$ & $ \textbf{7.1\%}$ \\
   
    \bottomrule
  \end{tabular}
\end{table}

\begin{figure}[ht!]
    \centering
    \includegraphics[width=0.7\textwidth]{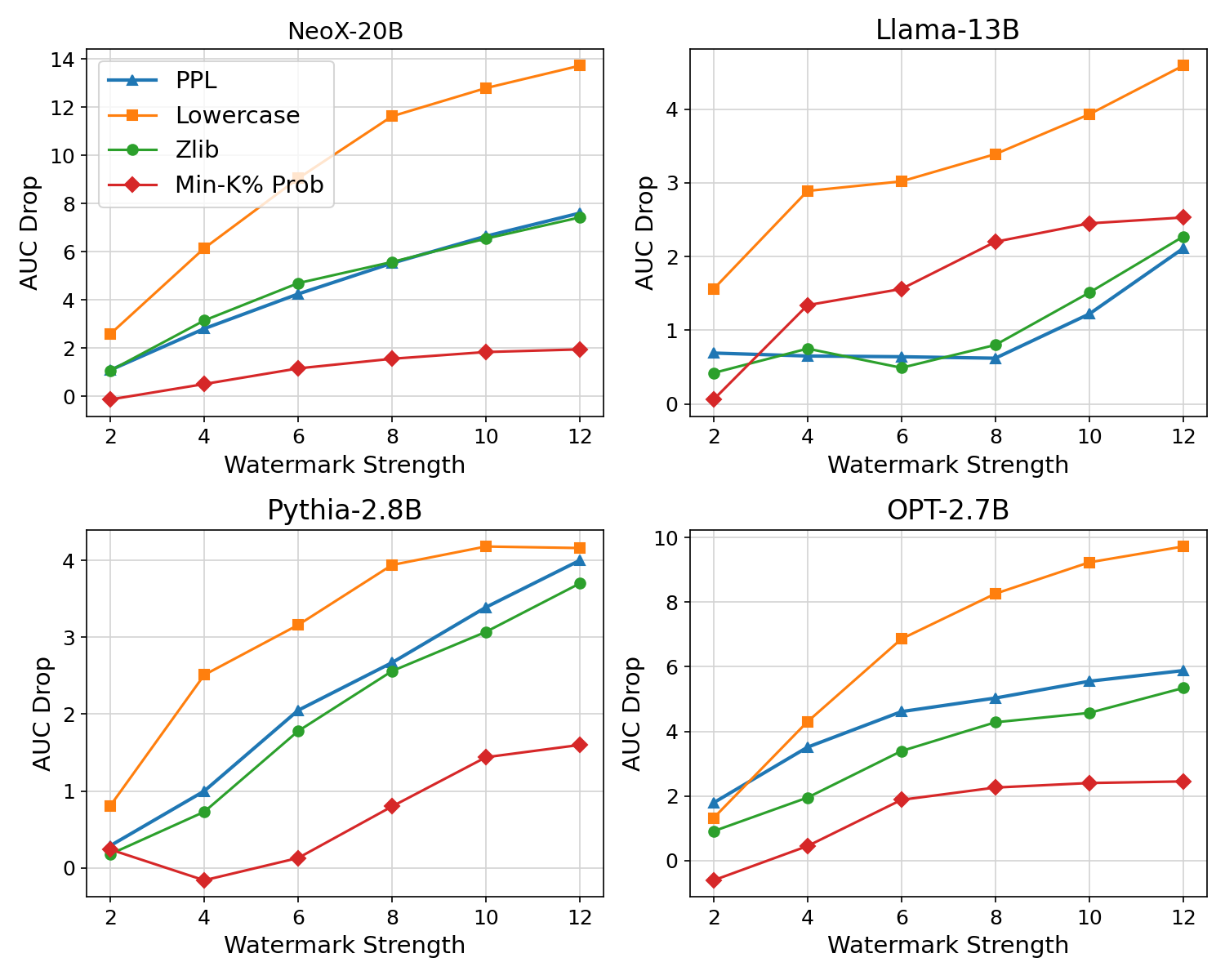}
    \caption{AUC drop due to watermarking for each MIA when varying the strength of the watermark.}
    \label{fig:mia_very_strength_appendix}
\end{figure}

\begin{table}[ht!]
    % \footnotesize
  \caption{We show the AUC drop when we vary the percentage of green tokens between $30\%$ and $70\%$. We bold the scenarios when a specific percentage value induces AUC drops for all the attacks.}
  \label{appendix-mia-vary-ratio}
  \centering
  \begin{tabular}{cc|ccccc}
    \toprule
    & & 0.3 & 0.4 & 0.5 & 0.6 & 0.7 \\

    \midrule
              & PPL & \textbf{0.71} & -0.10 & \textbf{1.40} & \textbf{2.14} & \textbf{1.65} \\
    Llama-30B & Lowercase & \textbf{0.27} & 2.60 & \textbf{4.24} & \textbf{2.45} & \textbf{3.15} \\
              & Zlib & \textbf{0.58} & -0.18 & \textbf{0.66} & \textbf{1.10} & \textbf{0.52} \\
              & Min-K\% Prob & \textbf{1.38} & 0.03 & \textbf{1.28} & \textbf{1.64} & \textbf{1.45} \\

        \midrule
              & PPL & \textbf{5.56} & \textbf{5.65} & \textbf{6.64} & \textbf{6.92} & \textbf{4.96} \\
    NeoX-20B & Lowercase & \textbf{9.84} & \textbf{11.12} & \textbf{12.79} & \textbf{11.94} & \textbf{9.78} \\
              & Zlib & \textbf{6.68} & \textbf{5.94} & \textbf{6.54} & \textbf{6.51} & \textbf{4.71} \\
              & Min-K\% Prob & \textbf{0.45} & \textbf{0.88} & \textbf{1.83} & \textbf{1.26} & \textbf{3.84} \\

        \midrule
              & PPL & \textbf{0.19} & -0.86 & \textbf{1.22} & \textbf{1.84} & \textbf{1.39} \\
    Llama-13B & Lowercase & \textbf{0.49} & 2.45 & \textbf{3.93} & \textbf{1.54} & \textbf{1.65} \\
              & Zlib & \textbf{1.31} & 0.28 & \textbf{1.51} & \textbf{2.00} & \textbf{1.29}\\
              & Min-K\% Prob & \textbf{2.81} & 1.21 & \textbf{2.45} &\textbf{ 2.70} & \textbf{2.78} \\

        \midrule
              & PPL & \textbf{4.65} & \textbf{4.49} & \textbf{3.39} & \textbf{4.42} & \textbf{3.66} \\
    Pythia-2.8B & Lowercase & \textbf{4.91} & \textbf{6.23} & \textbf{4.18} & \textbf{5.33} & \textbf{7.22} \\
              & Zlib & \textbf{5.37} & \textbf{3.97} & \textbf{3.07} & \textbf{3.17} & \textbf{2.20} \\
              & Min-K\% Prob & \textbf{1.10} & \textbf{1.21} & \textbf{1.44} & \textbf{3.42} & \textbf{4.71} \\

        \midrule
              & PPL & \textbf{5.45} & \textbf{5.39} & \textbf{5.55} & \textbf{5.18} & \textbf{5.76} \\
    OPT-2.7B & Lowercase & \textbf{7.71} & \textbf{9.91} & \textbf{9.23} & \textbf{9.83} & \textbf{7.28}\\
              & Zlib & \textbf{3.35} & \textbf{4.12} & \textbf{4.57} & \textbf{4.17} & \textbf{4.11} \\
              & Min-K\% Prob & \textbf{2.30} & \textbf{2.07} & \textbf{2.40} & \textbf{3.90} & \textbf{5.00} \\
   
    \bottomrule
  \end{tabular}
\end{table}

\clearpage
\newpage
\subsection{Additional experiments on BookMIA}

\label{bookmia-appendix}

In this section, we conduct experiments using models finetuned on a subset of BookMIA, which we refer to as BookMIA-2. To build BookMIA-2, we first select only the samples that were not part of the training set of any model that we consider (labeled as 0 by~\cite{shi2023detecting}). Then, we randomly select half of them as finetuning data (referred to as \textit{seen} samples) and keep the other half as \textit{unseen} samples. Note that for BookMIA-2, there would not be a distribution difference between the seen and unseen samples~\citep{das2024blind}. We also consider duplicating a sample from the training set of BookMIA-2 to have more fine-grained control over the memorization of that sample. We use Llama-7B in all the experiments from this section. \\ \\
\topic{Verbatim Memorization} We study verbatim memorization on BookMIA-2 by measuring the relative increase in perplexity on the generation of the duplicated sample by the watermarked model compared to the original model, as well as the ratio between the probability of generating the duplicated sample by the original model to the watermarked model, which we refer to as probability reduction factor. We run each experiment with 20 seeds and report the average perplexity and the minimum probability reduction factor. We consider both the UMD and Unigram watermarking methods with several strengths (2, 5, and 10) and prompt the model with an empty string, as well as with the first 10, 20, and 100 words from the training sample. Additionally, we consider several duplication factors (the number of times one randomly chosen target sample appears in the dataset): 1, 10, 20, and 50. We show the results in Table~\ref{verbatim-bookmia-finetuned}. We observe that even in high memorization cases (duplication factor of 50), as long as the watermark is strong enough, the probability of generating the memorized sample decreases by almost 200 orders of magnitude, making it very unlikely to be generated.

\begin{table}[ht!]
    % \scriptsize
  \caption{Average relative increase in perplexity and minimum probability reduction factor for generating the memorized target sample from BookMIA-2. Note that ``S.'', ``P.'', and ``D.'' stand for watermark method's strength, prompt length, and duplication factor, respectively.}
  \label{verbatim-bookmia-finetuned}
  \centering
  \begin{tabular}{c|c|c|cc|cc|cc|cc}
    \toprule
    & & & \multicolumn{2}{c|}{D = 1} & \multicolumn{2}{c|}{D = 10} & \multicolumn{2}{c|}{D = 20} & \multicolumn{2}{c}{D = 50} \\
    \midrule
      & S. & P. & PPL.  & Prob. & PPL. & Prob. & PPL. & Prob. & PPL. & Prob. \\
    \midrule

     & & 0 & 0.32 & $3.1 \times 10^{70}$ & 0.28 & $2.9 \times 10^{45}$ & 0.17 & $1.1 \times 10^{19}$ & 0.007 & $4.4 \times 10^{0}$  \\
     & 2 & 10 & 0.32 & $1.9 \times 10^{69}$ & 0.28 & $1.2 \times 10^{44}$ & 0.17 & $1.3 \times 10^{18}$ & 0.005 & $5.1 \times 10^{0}$  \\
     & & 20 & 0.32 & $1.7 \times 10^{67}$ & 0.28 & $2.9 \times 10^{42}$ & 0.17 & $4.8 \times 10^{17}$ & 0.005 & $5.0 \times 10^{0}$  \\
     & & 100 & 0.32 & $3.7 \times 10^{57}$ & 0.28 & $1.6 \times 10^{34}$ & 0.16 & $3.3 \times 10^{10}$ & 0.005 & $4.2 \times 10^{0}$  \\
    
    \cmidrule{2-11}
    
    & & 0 & 2.93 & $1.1 \times 10^{366}$ & 2.58 & $1.8 \times 10^{261}$ & 1.53 & $1.6 \times 10^{122}$ & 0.10 & $1.8 \times 10^{18}$ \\
   UMD & 5 & 10 & 2.92 & $7.6 \times 10^{360}$ & 2.56 & $6.9 \times 10^{257}$ & 1.51 & $8.9 \times 10^{99}$ & 0.09 & $1.1 \times 10^{16}$ \\
     & & 20 & 2.91 & $1.2 \times 10^{354}$ & 2.55 & $2.6 \times 10^{254}$ & 1.50 & $3.0 \times 10^{98}$ & 0.09 & $4.3 \times 10^{15}$ \\
     & & 100 & 2.89 & $4.9 \times 10^{313}$ & 2.53 & $4.2 \times 10^{213}$ & 1.45 & $6.0 \times 10^{71}$ & 0.09 & $6.7 \times 10^{13}$ \\

    \cmidrule{2-11}

     & & 0 & 38.6 & $3.2 \times 10^{1028}$ & 33.0 & $6.7 \times 10^{883}$ & 18.6 & $1.9 \times 10^{508}$ & 1.86 & $7.1 \times 10^{245}$  \\
     & 10 & 10 & 38.5 & $1.1 \times 10^{1006}$ & 32.8 & $9.2 \times 10^{869}$ & 18.4 & $2.5 \times 10^{495}$ & 1.78 & $4.7 \times 10^{226}$  \\
     & & 20 & 38.4 & $2.4 \times 10^{982}$ & 32.7 & $3.7 \times 10^{851}$ & 18.2 & $3.8 \times 10^{482}$ & 1.76 & $4.6 \times 10^{223}$  \\
     & & 100 & 38.0 & $2.8 \times 10^{860}$ & 32.3 & $6.6 \times 10^{731}$ & 17.7 & $1.0 \times 10^{388}$ & 1.70 & $3.8 \times 10^{199}$  \\

    \midrule

     & & 0 & 0.32 & $ 5.2 \times 10^{63}$ & 0.29 & $ 6.8 \times 10^{59}$ & 0.17 & $ 1.2 \times 10^{14}$ & 0.008 & $ 2.9 \times 10^{0}$ \\
     & 2 & 10 & 0.32 & $ 4.7 \times 10^{62}$ & 0.29 & $ 3.6 \times 10^{58}$ & 0.17 & $ 4.8 \times 10^{14}$ & 0.005 & $ 5.9 \times 10^{0}$ \\
     & & 20 & 0.32 & $ 7.9 \times 10^{61}$ & 0.29 & $ 9.8 \times 10^{56}$ & 0.17 & $ 3.1 \times 10^{14}$ & 0.005 & $ 5.8 \times 10^{0}$ \\
     & & 100 & 0.31 & $ 1.2 \times 10^{50}$ & 0.28 & $ 5.2 \times 10^{46}$ & 0.16 & $ 2.7 \times 10^{7}$ & 0.005 & $ 5.1 \times 10^{0}$ \\
    
    \cmidrule{2-11}
    
    & & 0 & 2.88 & $ 3.4 \times 10^{304}$ & 2.56 & $ 1.2 \times 10^{290}$ & 1.53 & $ 1.1 \times 10^{79}$ & 0.11 & $ 2.2 \times 10^{22}$ \\
   Unigram & 5 & 10 & 2.87 & $ 1.6 \times 10^{300}$ & 2.56 & $ 1.9 \times 10^{286}$ & 1.52 & $ 8.6\times 10^{77}$ & 0.10 & $ 5.9 \times 10^{16}$ \\
     & & 20 & 2.87 & $ 8.2 \times 10^{295}$ & 2.55 & $ 3.1 \times 10^{282}$ & 1.50 & $ 6.2\times 10^{76}$ & 0.09 & $ 4.6 \times 10^{16}$ \\
     & & 100 & 2.82 & $ 2.5 \times 10^{261}$ & 2.50 & $ 6.6 \times 10^{239}$ & 1.45 & $ 1.0 \times 10^{47}$ & 0.09 & $ 4.1 \times 10^{14}$ \\

    \cmidrule{2-11}

     & & 0 & 37.6 & $ 2.4 \times 10^{834}$ & 32.3 & $ 2.2 \times 10^{811}$ & 18.5 & $ 1.8 \times 10^{425}$ & 1.87 & $ 2.6 \times 10^{268}$ \\
     & 10 & 10 & 37.7 & $ 3.5 \times 10^{824}$ & 32.3 & $ 7.1 \times 10^{799}$ & 18.3 & $ 1.3 \times 10^{419}$ & 1.80 & $ 5.7 \times 10^{251}$ \\
     & & 20 & 37.6 & $ 4.9 \times 10^{812}$ & 32.2 & $ 2.3 \times 10^{788}$ & 18.1 & $ 6.1 \times 10^{405}$ & 1.78 & $ 1.3 \times 10^{248}$ \\
     & & 100 & 36.9 & $ 2.4 \times 10^{707}$ & 31.5 & $ 9.5 \times 10^{684}$ & 17.4 & $ 1.5 \times 10^{323}$ & 1.72 & $ 1.0 \times 10^{212}$ \\
      
    \bottomrule
  \end{tabular}
\end{table}

\topic{MIA} We also study the effect of the watermark on the effectiveness of MIAs for copyrighted training data detection (on BookMIA-2, without duplicated samples). We show the results in Table~\ref{main-bookmia-finetuned} and observe that watermarking negatively affects MIAs' success rate, which is consistent with our previous findings (from Section~\ref{mia}). Finally, we also run our adaptive method from Section~\ref{adaptive} and observe an improvement of $0.9\%$ over Min-K\% Prob.
~\begin{figure}[ht!]
    \centering
    \includegraphics[width=0.8\textwidth]{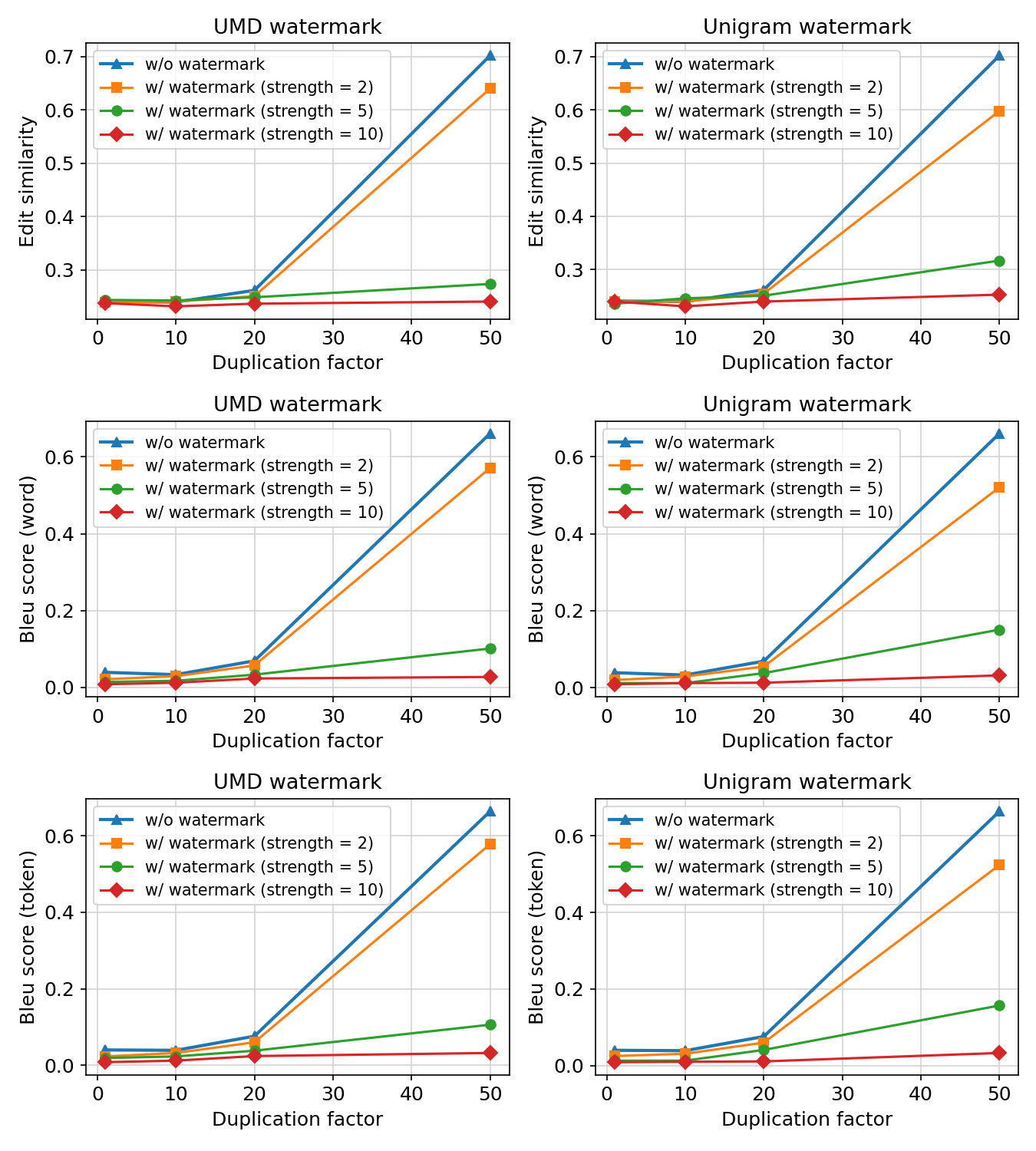}
    \caption{Edit similarity (\textit{top}), word-level BLEU score (\textit{middle}), and token-level BLEU score (\textit{bottom}) between the generated completion and the ground truth when considering different watermark strengths on BookMIA-2.}
    \label{fig:approximate_mem}
\end{figure}

\begin{table}[ht!]
% \scriptsize
\caption{AUC of each MIA for the unwatermarked (\textit{top} of each cell), watermarked models (\textit{middle} of each cell), and the drop between the two (\textit{bottom} of each cell) on BookMIA-2 (without any duplicated samples) using the UMD scheme with a strength of 10. We average the results over 5 runs with different seeds.}  \label{main-bookmia-finetuned}
  \centering
  \begin{tabular}{c|c}
    \toprule
     & Llama-7B (fine-tuned)  \\
    \midrule
       & $ 58.5 \pm 0.0\% $ \\
   PPL & $ 56.6 \pm 0.0\%  $ \\
       & \textbf{ 1.9\%} \\
    \midrule
             & $ 59.8 \pm 0.1\% $ \\
   Lowercase & $ 52.9 \pm 0.3\%  $ \\
             & \textbf{ 6.9\%} \\
    \midrule
        & $ 59.7 \pm 0.0\% $ \\
   Zlib & $ 56.1 \pm 0.1\% $ \\
        & \textbf{ 3.6\%} \\
    \midrule
             & $ 58.5 \pm 0.0 \% $ \\
Min-K\% Prob & $ 57.1 \pm 0.2\% $ \\
             & \textbf{ 1.4\%} \\
   
    \bottomrule
  \end{tabular}
\end{table}

\subsection{Theoretical analysis}

\label{theoretical}

\topic{Notations and assumptions} We assume that the set of all copyrighted texts $C_D$ that were part of the training data has $m$ elements $\{ s_1, s_2, ..., s_m \}$. Also, we assume that each copyrighted text has a fixed length $n$, and they are independent from each other.

\begin{theorem}\label{theorem1}
        For an LLM watermarked using a ``hard'' UMD scheme with a percentage of $\gamma$ green tokens, then the probability of generating a copyrighted text is lower than $m \cdot \gamma^n$.
\end{theorem}

\topic{Proof} Given one sample $s = t_1 \oplus t_2 \oplus ... \oplus t_n \in C_D$. For a ``hard'' watermarking scheme, the probability $P(s)$ of generating $s$ is smaller than the probability of each token $t_i$ to be on a green list. So, $P(s) < \gamma^n$. The probability of not generating any $s_j \in C_D$ is $P(\neg s_1 \wedge \neg s_2 \wedge ... \wedge \neg s_m) = \prod_{i=\overline{1, m}}(1 - P(s_i)) > (1-\gamma^n)^m > 1 - m \gamma^n.$ Note that we used Bernoulli's inequality at the end. So, the probability of generating at least one copyrighted text is lower than $1 - (1 - m \cdot \gamma^n)$ and hence lower than $m \cdot \gamma^n$. \\ \\

\topic{Example} Let's consider a ``hard'' UMD watermarking scheme with $\gamma = 0.5$. Let's assume each copyrighted text is $100$ tokens, the model was trained on a dataset containing $10^9$ copyrighted texts. The probability to generate a copyrighted text is $ < 10^9 \cdot 0.5^{100} = \frac{10^{9}}{2^{100}} = \frac{1000^3}{1024^{10}} < \frac{1000^3}{1000^{10}} = 1000^{-7} = 10^{-21}$ and hence very low.

\begin{theorem}\label{theorem2}
    Let $f$ be a LLM and $f_W$ its watermarked version with a ``soft'' UMD scheme and let $\epsilon \in (0, \frac{1}{4})$. Let $s = t_1 \oplus t_2 \oplus ... \oplus t_n \in C_D$ be a copyrighted sample. We consider $\gamma = 0.5$. We denote the output of the softmax layer of $f$ for generating the token $t_i$ as $\frac{a_i}{d_i + a_i}$ and in the case of $f_W$, we denote it by $\frac{a_i \cdot e^{\delta}}{b_i^{'} + c_i^{'} \cdot e^{\delta} + a_i \cdot e^{\delta}}$ (if $t_i$ is on the green list) and $\frac{a_i}{b_i^{''} + c_i^{''} \cdot e^{\delta} + a_i}$ (if $t_i$ is on the red list), where $a_i$ is the exponential of the logit value corresponding to the token $t_i$ and $b_i^{'},b_i^{''}$ and $c_i^{'}, c_i^{''}$ are the sum of the exponentials of the logits corresponding to other tokens that are on the red list and green list, respectively. We assume that $\frac{x}{a_i} < M = \frac{1-4\epsilon}{1+4\epsilon}$, for all $x \in \{ d_i, b_i^{'}, b_i^{''}, c_i^{'}, c_i^{''} \}$ which would restrict $f$ to be relatively confident in its predictions for each token $t_i$. Then, we can always find a $\delta$ (strength) for the watermarking scheme such that the probability of generating $s$ is reduced by at least $(1+\frac{2\epsilon}{2\epsilon+1})^n$ times in comparison to the case of the unwatermarked model.
\end{theorem}

\topic{Proof} First, we observe that the probability of generating the token $t_i$ by the unwatermarked model is $\frac{a_i}{d_i+a_i} = \frac{1}{\frac{d_i}{a_i} + 1} > \frac{1}{M + 1} = 1/2 + 2\epsilon$. \\ \\
We observe that since there is a finite number of $\frac{x}{a_i}'s$ and they are all positive, then it exist a lower bound for $\frac{x}{a_i}$ (let's denote it by $m > 0$). Since $\gamma = 0.5$, the probability of $t_i$ being a green token is $\frac{1}{2}$ and hence the probability of the watermarked model to generate $t_i$ is $\frac{1}{2} \frac{a_i \cdot e^{\delta}}{b_i^{'} + c_i^{'} \cdot e^{\delta} + a_i \cdot e^{\delta}} + \frac{1}{2} \frac{a_i}{b_i^{''} + c_i^{''} \cdot e^{\delta} + a_i} < \frac{1}{2} + \frac{1}{2} \frac{a_i}{b_i^{''} + c_i^{''} \cdot e^{\delta} + a_i} = \frac{1}{2} + \frac{1}{2} \frac{1}{\frac{b_i^{''}}{a_i} + \frac{c_i^{''}}{a_i} \cdot e^{\delta} + 1} \leq \frac{1}{2} + \frac{1}{2} \frac{1}{m \cdot (e^{\delta}+1) + 1}$. We pick $\delta > log\big( \frac{1-2\epsilon (m+1)}{2\epsilon m} \big)$ and we observe that $\frac{1}{2} + \frac{1}{2} \frac{1}{m \cdot (e^{\delta}+1) + 1} < \frac{1}{2} + \frac{1}{2} \frac{1}{m \cdot (\frac{1-2\epsilon (m+1)}{2\epsilon m}+1) + 1} = \frac{1}{2} + \frac{1}{2} \frac{1}{m \cdot (\frac{1-2\epsilon}{2\epsilon m}) + 1} = \frac{1}{2} + \epsilon$. \\ \\
So, by combining the two observations above, we conclude that the probability of generating $t_i$ is reduced by at least $\frac{\frac{1}{2}+2\epsilon}{\frac{1}{2}+\epsilon} = 1 + \frac{2\epsilon}{2\epsilon + 1}$ times. Therefore, since there are $n$ tokens in $s$, the probability of generating $s$ is reduced by at least $(1 + \frac{2\epsilon}{2\epsilon + 1})^n$ times.

\topic{Observation} Since the probability is reduced by at least $(1 + \frac{2\epsilon}{2\epsilon + 1})^n$ times in Theorem~\ref{theorem2} then the probability of generating $s$ is lower than $(\frac{2\epsilon+1}{4\epsilon + 1})^n$ (as the maximum probability of generating with the unwatermarked model is $1$). Hence, as in Theorem~\ref{theorem1}, we observe that the probability of generating a copyrighted text is lower than $m \cdot (\frac{2\epsilon+1}{4\epsilon + 1})^n$. \\ \\
\topic{Takeaways} Our theoretical analysis demonstrates that watermarking significantly reduces the probability of generating copyrighted text verbatim. For both a ``hard'' and ``soft'' UMD scheme, the upper bound for the likelihood of producing copyrighted content decreases exponentially with the length of the copyrighted texts.

\subsection{Metrics used in MIAs}

In this section we provide short formulas for several metrics used by MIAs such as Smaller Ref, Lowercase and Zlib~\citep{carlini2021extracting}. \\ \\ 
Let $f$ be the model the MIA is applied to, $g$ be a smaller model, $x$ be a sample (string), $\texttt{log}$ be the natural logarithm function, $\texttt{perplexity\_h(x)}$ be a function that computes the perplexity of a model $h$ on $x$, $\texttt{lowercase}$ be a function that maps a string to its lowercased version, and $\texttt{zlib(x)}$ be a function that computes the zlib entropy of a string $x$. \\ \\
Smaller Ref uses the metric $\frac{\texttt{log(\texttt{perplexity\_f(x)})}}{\texttt{log(\texttt{perplexity\_g(x)})}}$. \\ \\
Lowercase uses the metric $\frac{\texttt{log(\texttt{perplexity\_f(x)})}}{\texttt{log(\texttt{perplexity\_f(\texttt{lowercase}(x))})}}$. \\ \\
Zlib uses the metric $\frac{\texttt{log(\texttt{perplexity\_f(x)})}}{\texttt{zlib}(x)}$. \\ \\

\subsection{Adaptive Min-K\% Prob}

\begin{table*}[ht!]
    % \footnotesize
  \caption{We show the AUC of Min-\%K Prob (referred as ``Not adapt.'') and our method (referred as ``Adapt.'') when using UMD watermarking scheme. We highlight the cases when our method improves over the baseline.}
  \label{ours-full}
  \centering
  \begin{tabular}{c|c|cccccc}
    \toprule
    & & Llama-30B & NeoX-20B & Llama-13B & Pythia-2.8B & OPT-2.7B \\
    \midrule

   WikiMIA    & Not adapt. & $66.2\%$ & $67.1\%$ & $64.5\%$ & $61.0\%$ & $55.7\%$ \\
   32 & Adapt. & $\textbf{68.5\%}$ & $\textbf{71.3\%}$ & $\textbf{66.3\%}$ & $61.0\%$ & $\textbf{59.1\%}$ \\

    \midrule
    WikiMIA   & Not adapt. & $64.4\%$ & $67.7\%$ & $62.8\%$ & $59.8\%$ & $55.3\%$ \\
   64 & Adapt. & $\textbf{67.3\%}$ & $\textbf{72.0\%}$ & $\textbf{64.9\%}$ & $\textbf{60.6\%}$ & $\textbf{57.4\%}$ \\

    \midrule
    WikiMIA   & Not adapt. & $70.0\%$ & $73.0\%$ & $68.9\%$ & $64.8\%$ & $59.2\%$ \\
   128 & Adapt. & $\textbf{73.1\%}$ & $\textbf{75.9\%}$ & $\textbf{71.0\%}$ & $\textbf{66.4\%}$ & $\textbf{64.0\%}$ \\

    \midrule
     WikiMIA  & Not adapt. & $70.5\%$ & $76.2\%$ & $70.4\%$ & $69.5\%$ & $63.1\%$ \\
   256 & Adapt. & $\textbf{71.3\%}$ & $\textbf{78.2\%}$ & $\textbf{72.4\%}$ & $\textbf{70.7}\%$ & $\textbf{66.2\%}$ \\

    \bottomrule
  \end{tabular}
\end{table*}

\subsection{Adaptive Zlib and Adaptive Lowercase}

We have also adapted Zlib, as well as Lowercase. In these adaptations, we use a method similar to Adaptive Min-K\% Prob to approximate pre-watermarked probabilities, then we apply Zlib / Lowercase. The results are presented in Tables~\ref{ours-full2} and~\ref{ours-full3}, where each double-cell shows the AUC score for the non-adaptive attack on top and the adaptive version on the bottom. We observe that adaptation improves the baseline in over 80\% of cases (for Zlib) and over 90\% of cases (for Lowercase). However, these adaptive methods appear slightly less effective compared to Min-K\% Prob. We hypothesize that this may be due to them using all token probabilities rather than the minimum K\% subset, as in Min-K\% Prob, which may increase error accumulation due to our token probability approximation.

\begin{table*}[ht!]
    % \footnotesize
  \caption{We show the AUC of Zlib (referred as ``Not adapt.'') and our method (referred as ``Adapt.'') when using UMD watermarking scheme. We highlight the cases when our method improves over the baseline.}
  \label{ours-full2}
  \centering
  \begin{tabular}{c|c|ccc}
    \toprule
    & & Llama-30B & NeoX-20B & Llama-13B \\
    \midrule

   WikiMIA    & Not adapt. & 64.4\% & 63.2\% & 62.3\% \\
   32 & Adapt. & \textbf{66.8\%} &	\textbf{64.2\%} 	& \textbf{64.9\%} \\

    \midrule
    WikiMIA   & Not adapt. & 62.4\% &	62.0\% & 	59.9\% \\
   64 & Adapt. & \textbf{66.8\%} &	\textbf{63.6\%} &	\textbf{64.8\%} \\

    \midrule
    WikiMIA   & Not adapt. & 68.6\% &	66.3\% &	65.8\% \\
   128 & Adapt. & \textbf{71.4\%} 	& \textbf{67.3\%} &	\textbf{69.1\%} \\

    \midrule
     WikiMIA  & Not adapt. & 72.0\% 	& 66.6\% 	& 71.6\% \\
   256 & Adapt. & 71.3\% 	& \textbf{67.1\%} &	71.1\% \\

    \bottomrule
  \end{tabular}
\end{table*}

\begin{table*}[ht!]
    % \footnotesize
  \caption{We show the AUC of Lowercase (referred as ``Not adapt.'') and our method (referred as ``Adapt.'') when using UMD watermarking scheme. We highlight the cases when our method improves over the baseline.}
  \label{ours-full3}
  \centering
  \begin{tabular}{c|c|ccc}
    \toprule
    & & Llama-30B & NeoX-20B & Llama-13B \\
    \midrule

   WikiMIA    & Not adapt. & 54.9\% & 59.4\% & 54.2\% \\
   32 & Adapt. & \textbf{57.4\%} &	\textbf{62.0\%} 	& \textbf{56.9\%} \\

    \midrule
    WikiMIA   & Not adapt. & 54.8\% &	56.8\% & 	53.8\% \\
   64 & Adapt. & \textbf{56.7\%} &	\textbf{59.4\%} &	\textbf{56.0\%} \\

    \midrule
    WikiMIA   & Not adapt. & 55.9\% &	58.2\% &	55.1\% \\
   128 & Adapt. & \textbf{57.2\%} 	& \textbf{59.4\%} &	\textbf{56.0\%} \\

    \midrule
     WikiMIA  & Not adapt. & 63.8\% 	& 55.4\% 	& 61.6\% \\
   256 & Adapt. & 63.5\% 	& \textbf{58.1\%} &	\textbf{62.3\%} \\

    \bottomrule
  \end{tabular}
\end{table*}

\subsection{Examples of generated samples}

Below we provide text samples generated from the unwatermarked model and from models with various UMD watermark strengths. All examples are generated from an empty string (free generation). These examples have relative perplexity increases that match or even exceed the corresponding averages in Figure~\ref{fig:verbatim_vary_strength}. \\

w/o watermark: ``Nonviolent communication, also called compassionate communication, is a way of relating to others based on values of human dignity, honesty, and empathy ...'' \\

w/ watermark (strength = 2): ``The Latifa Hospital was established in Dubai, United Arab Emirates in March 1983 to provide advanced health care within Dubai ...'' \\

w/ watermark (strength = 4): ``We all make New Year's Resolutions and all of them start with “getting fitter” :) If it happens to be in your list, let me know ...'' \\

w/ watermark (strength = 6): ``I'll be the first to say that MTS doesn't always achieve what it sets out to do at first look but you can see the software is reaching its potential ...'' \\

w/ watermark (strength = 8): ``United Nations : Member states of UNO climate change conference agreed on Thursday that progress had to accelerate in implementing ...'' \\

w/ watermark (strength = 10): ``Like our Facebook page to see when the weekend is set and to find out about local motorcycling news, events and ...'' \\

w/ watermark (strength = 12): ``Your child will learn time management as well: it will learn how much time to spend doing each assignment and studying for tests to ensure they get them done ...'' \\

w/ watermark (strength = 14): ``Grant is a shy four to six year old beagle boy with one blind eye and lacking much vision in both eyes. His family had been unable to ...'' \\

w/ watermark (strength = 16): ``We are leading online doctor shopping solution company in Noida. It offers the best web solutions with goal orientated expertise in medical sector website development and marketing ...''

\subsection{Extended Related Work}

\topic{Memorization}
One cause of copyright issues is that machine learning models may memorize training data. 
Prior studies have observed that LLMs can memorize private information in training data, such as phone numbers and addresses \citep{karamolegkou2023copyright, carlini2019secret, carlini2021extracting, lee2021deduplicating}, leading to significant privacy and security concerns. 
To measure memorization, \citet{carlini2021extracting} proposes eidetic memorization, defining a string as memorized if it was present in the training data and it can be reproduced by a prompt. 
This definition, along with variations like exact and perfect memorization, has been widely adopted in subsequent studies \citep{tirumala2022memorization, kandpal2022deduplicating}.
\citet{carlini2022quantifying} quantitatively measures memorization in LLMs as the fraction of extractable training data and finds that memorization significantly grows as model size scales and training examples are duplicated.
To minimize memorization, \citet{lee2021deduplicating} and \citet{kandpal2022deduplicating} propose deduplicating training data, which also improves accuracy. \citet{hans2024like} proposes the Goldfish Loss as a training-time defense against verbatim memorization.
\citet{ippolito2023preventing} proposes an inference time defense that perfectly prevents all verbatim memorization. However, it cannot prevent the leakage of training data due to the existence of many ``style-transfer'' prompts, suggesting it is a challenging open problem. Unlike the methods that we are studying in this paper, \citet{ippolito2023preventing} requires access to a complete set of copyrighted texts that the model was trained on.
Memorization in the image domain has also been studied from various angles \citep{somepalli2023diffusion, somepalli2023understanding, carlini2023extracting, wen2024detecting}. \\ \\
\topic{Membership Inference}
As a proxy for measuring memorization, membership inference attacks (MIAs) predict whether or not a particular example was used to train the model \citep{shokri2017membership, yeom2018privacy, bentley2020quantifying}. 
Most membership inference attacks rely only on the model’s
loss since the model is more likely to overfit an example if it is in the training data \citep{sablayrolles2019white}. 
\citet{carlini2022membership} trains shadow models to predict whether an example is from the training data.
In the NLP domain, many works have focused on masked language models \citep{mireshghallah2022quantifying} and fine-tuning data detection \citep{song2019auditing, shejwalkar2021membership}.
Recently, \citet{shi2023detecting} studies pretraining data inference and introduced a detection method based on the hypothesis that unseen examples are likely to contain outlier words with low probabilities under the LLM. 
\citet{zhang2024min} approaches pretraining data detection by measuring how sharply peaked the likelihood is around the inputs. 
\citet{duarte2024cop} proposes detecting copyrighted content in training data by probing the LLM with multiple-choice questions, whose options include both verbatim text and their paraphrases.
Other methods include testing perplexity differences \citep{mattern-etal-2023-membership} and providing provable guarantees of test set contamination without access to pretraining data or model weights \citep{oren2023proving}.

\subsection{Computing Infrastructure} All of our experiments were run on either three Nvidia RTX A6000 or four Nvidia RTX A5000 GPUs, using 128 GB of memory. We used the Transformers library (version 4.35.2) and PyTorch (version 2.1.0).

\subsection{Limitations \& Discussion} We emphasize that our claim is not that any watermarking method inherently prevents the generation of verbatim copyrighted text. Instead, we demonstrate that popular watermarking methods can unintentionally produce this effect. We believe that a distortion-free watermark, such as the one proposed by~\citet{kuditipudi2023robust}, would not have this unintended effect. Our proposed method for improving MIAs' success rate on watermarked models makes strong assumptions on the watermarking scheme, which may not always be satisfied despite empirical improvements in our experiments. Our observations on the deterioration of MIAs' success suggests that for copyright violation auditing, an unwatermarked model or the watermarking scheme may be needed. We encourage the community to further refine adaptive methods to ensure robust copyright protection and data privacy, and consider the interactions of different methods on downstream legal concerns.